\documentclass[conference,compsoc]{IEEEtran}

\pdfoutput=1

\usepackage[pdftex]{graphicx}
\usepackage{array,booktabs,ragged2e}
\usepackage[tight,footnotesize]{subfigure}
\usepackage{pbox}
\usepackage{amsmath}
\usepackage{amsfonts}
\usepackage{epsfig}
\usepackage{url}
\usepackage{multirow}
\usepackage[ruled,vlined,linesnumbered]{algorithm2e}
\usepackage{setspace}
\usepackage{wrapfig}
\usepackage{comment}
\usepackage{algorithmic}
\usepackage{tablefootnote}
\usepackage{hyperref}
\usepackage{float}
\usepackage{bibspacing}
\usepackage{balance}

\usepackage{xcolor}
\hypersetup{
	colorlinks,
	linkcolor={red},
	citecolor={black},
	urlcolor={blue}
}

\newcommand{\etal}{{\em et al.}}

\newcommand{\bit}{\begin{itemize}}
	\newcommand{\eit}{\end{itemize}}
\newcommand{\ben}{\begin{enumerate}}
	\newcommand{\een}{\end{enumerate}}
\newcommand{\beq}{\begin{equation}}
\newcommand{\eeq}{\end{equation}}

\newcommand{\lcare}{{\normalsize \textsf{CARE}}} 
\newcommand{\care}{{\small \textsf{CARE}}} 
\newcommand{\scare}{{\footnotesize \textsf{CARE}}} 
 
\newcommand{\avgknn}{{\small \textsf{AvgKNN}}} 
\newcommand{\savgknn}{{\footnotesize \textsf{AvgKNN}}} 
\newcommand{\lof}{{\small \textsf{LOF}}} 
\newcommand{\slof}{{\footnotesize \textsf{LOF}}} 
\newcommand{\fvps}{{\small \textsf{FVPS}}}

\ifCLASSOPTIONcompsoc
  \usepackage[nocompress]{cite}
\else
  \usepackage{cite}
\fi

\begin{document}
\title{Sequential Ensemble Learning for Outlier Detection: A Bias-Variance Perspective}


\author{\IEEEauthorblockN{Shebuti Rayana \quad\quad\quad\quad Wen Zhong \quad\quad\quad\quad Leman Akoglu} 
\IEEEauthorblockA{Department of Computer Science \\
Stony Brook University, Stony Brook, NY 11794\\
Email: \{srayana,wzhong,leman\}@cs.stonybrook.edu}}

\maketitle

\begin{abstract}
Ensemble methods for classification and clustering have been effectively used for decades, while ensemble learning for outlier detection has only been studied recently. In this work, we design a new ensemble approach for outlier detection in multi-dimensional point data, which provides improved accuracy by reducing error through both bias and variance. Although classification and outlier detection appear as different problems, their theoretical underpinnings are quite similar in terms of the bias-variance trade-off~\cite{aggarwal2015theoretical}, where outlier detection is considered as a binary classification task with unobserved labels but a similar bias-variance decomposition of error.

In this paper, we propose a sequential ensemble approach called \care~that employs a two-phase aggregation of the intermediate results in each iteration to reach the final outcome. Unlike existing outlier ensembles which solely incorporate a parallel framework by aggregating the outcomes of independent base detectors to reduce variance, our ensemble incorporates both the parallel and sequential building blocks to reduce bias as well as variance by 
($i$) successively eliminating outliers from the original dataset to build a better data model on which outlierness is estimated (sequentially), and ($ii$) combining the results from individual base detectors and across iterations (parallelly).
Through extensive  experiments on sixteen real-world datasets mainly from the UCI machine learning repository~\cite{Lichman:2013}, we show that \care~performs significantly better than or at least similar to the individual baselines. We also compare \care~with the state-of-the-art outlier ensembles where it also provides significant improvement when it is the winner and remains close otherwise.  
\end{abstract}

\IEEEpeerreviewmaketitle

\section{Introduction}
\label{sec:intro}
As a significant subject, outlier detection is widely researched in the literature. There exist various approaches for outlier detection such as density based methods~\cite{breunig2000lof,kriegel2009loop,papadimitriou2003loci} and distance based methods \cite{zhang2009new, otey2006fast}, which find unusual points by the distance to their $k$ nearest neighbors (kNNs). However, each of these methods can only focus on some specific kinds of outliers based on the datasets collected from different application domains. There exists no known algorithm that could detect all types of outliers that appear in a wide variety of domains. As a result, ensemble learning for outlier detection has become a popular research area more recently \cite{zimek2013subsampling,rayana2015less,aggarwal2015theoretical}, which aims to put together multiple detectors so as to leverage the ``strength of the many".

In contrast to outlier detection ensembles, classification ensembles have been studied for decades.
The explosive growth of classification ensemble models provides the new opportunity to design effective methods for other machine learning tasks including outlier mining. One can categorize ensemble methods into two kinds. The first one is the parallel ensemble, where base learners are created independent of each other and their results are combined to get the final outcome; while the second one is the sequential ensemble, where base learners are created over iterations and have dependency among them. Specifically, several outlier ensembles are proposed  based on two seminal works of classification ensembles: ($i$) the parallel ensemble Bagging \cite{brei96}, which creates base components from different subsamples of training datasets parallelly, and ($ii$) the sequential ensemble AdaBoost \cite{freund1997decision}, which creates base components iteratively. Among those, some try to induce diversity among the base detectors \cite{lazarevic2005feature, zimek2013subsampling, aggarwal2015theoretical}, and others selectively combine outcomes from the candidate detectors \cite{rayana2015less,gao2006converting}.

Existing outlier ensembles have several limitations, most importantly they avoid discussing the theoretical aspects of outlier detection. Recently, Aggarwal \etal~\cite{aggarwal2015theoretical} argue that although they appear to be very different problems, classification and outlier detection share quite similar theoretical underpinnings in terms of the bias-variance trade-off. Specifically, one can consider the outlier detection problem as a binary classification task where the labels are unobserved, the inliers being the majority class and the outliers the minority class, and the error of a detector can be decomposed into bias and variance terms in a similar way.  In existing outlier ensembles, various parallel frameworks combining multiple detector outcomes are designed to reduce variance only,  most of which are incapable of overcoming the presence of inaccurate base detectors. On the other hand, it remains challenging to reduce bias in a controlled way for outlier detection or remove inaccurate detectors due to the lack of ground truth to validate the results during the intermediate steps. There exist some successful heuristic approaches to reduce bias. One such commonly used approach is to remove outliers in successive iterations~\cite{journals/sigkdd/Aggarwal12} to build more robust outlier models iteratively.

In this paper, we study the feasibility of bias-variance reduction under the unsupervised setting, and propose a sequential ensemble model called Cumulative Agreement Rates Ensemble (\care), to reduce both bias and variance for outlier detection. Specifically, each iteration in the sequential ensemble consists of two aggregation phases: (1) in the first phase, we combine the results of feature-bagged base detectors using weighted aggregation, where weights are estimated in an unsupervised way through the Agreement Rates (AR) method by~\cite{platanios2014estimating}, and (2) in the second phase, the result of the current iteration is aggregated with the combined result from the previous iterations cumulatively. These two phase aggregations in each iteration aim to reduce the variance. Furthermore, we use the combined result from the previous iterations to improve the next iteration by removing the top (i.e., most obvious) outliers and perform a variable probability sampling to create the data model to be used for the next iteration. The removal of top outliers in successive iterations aims to reduce the bias.

To the best of our knowledge, this is the first work focusing on reducing both bias and variance for unsupervised outlier detection. In general, this paper offers the following contributions: \\
\vspace{-0.15in}
\begin{itemize}
\setlength{\itemsep}{-0.95\itemsep}
\item We design a new approach which incorporates weighted aggregation of feature-bagged base detectors, where weights are estimated in an unsupervised fashion (Section~\ref{sssec:error} and~\ref{sssec:weight}).
\item We devise a sequential ensemble over the weighted combination, which cumulatively aggregates the results from multiple iterations until a stopping condition is met (Section~\ref{sssec:seqEn} and~\ref{sssec:SC}).
\item We provide a new sampling approach called Filtered Variable Probability Sampling (\fvps) which utilizes the result from the previous iteration to filter the top outliers, and uses variable probability sampling to select points from the original data to create the data model for the next iteration (Section~\ref{sssec:seqEn}).   
\item Our sequential ensemble is designed to reduce both bias and variance and improves the overall result. Moreover, we provide experiments with synthetic datasets to support this claim (Section~\ref{sec:bv}).
\end{itemize}
We evaluate our method on sixteen different real-world datasets majority of which are from the UCI machine learning repository~\cite{Lichman:2013}. Our results show that \care~outperforms the baseline (i.e., non-ensemble) detectors in most cases  and  remains close to the baselines in cases where it falls shorter. We also compare \care~with the  existing state-of-the-art outlier ensembles~\cite{zimek2013subsampling, aggarwal2015theoretical, liu2008isolation}. Similarly, it provides significant improvement when it is the winner, and performs close otherwise. (Section \ref{sec:eval})

\section{Related Work}
\label{sec:related}

\vspace{-0.05in}
\subsection{Ensemble Models}
\vspace{-0.1in}
Ensemble models for classification have been extensively studied in the literature for several decades. In 1996, Breiman~\cite{brei96} presented a parallel ensemble, today well-known as \textit{bagging}, which consists of multiple predictor components trained on samples of the original dataset, to infer the label using a plurality vote, enhancing the model through variance reduction. However, bagging omitted the bias term and could only reduce the variance. To make up this deficit, a new sequential ensemble known as \textit{boosting} was devised by Freund \etal~\cite{freund1997decision}. Their proposed AdaBoost algorithm assigned larger weights for the misclassified instances to advance the given base classification algorithm and combined weighted sum of multiple weak learners into a boosted classifier to reduce both bias and variance.

After these two important seminal works, a proliferation of ensemble methods followed, aiming to explain and improve over the original methods. A two-stage process called adaptive bagging \cite{breiman1999using} was proposed to perform bias and variance reduction respectively. It generated an intermediate output in the first stage and the altered first stage output was adopted as new input of the second stage that is bagging. Sun \etal~\cite{sun2007cost} analyzed the influence of adding a cost term to AdaBoost and offered the cost-sensitive boosting algorithm for imbalanced data classification.

Our proposed outlier detection approach \care~uses similar insights as in AdaBoost; by sequentially updating the data model on which data points are scored for outlierness as well as by combining multiple detector outcomes parallelly to reduce bias and variance, respectively.
Since the above methods can be learned in supervised settings while most anomaly detection tasks provide no labels, our method differs from the existing classifier ensembles in that we focus on reducing both bias and variance in a fully unsupervised setting, which has not been studied before.

\vspace{-0.05in}
\subsection{Outlier Ensembles}
\vspace{-0.1in}
Outlier ensemble learning, as a rarely explored area, mainly tries to reduce the variance through the combination of different base detectors. A parallel approach called feature bagging, proposed by Lazarevic and Kumar~\cite{lazarevic2005feature}, built an ensemble based on randomly selected feature subsets from original features to detect outliers in high-dimensional and noisy datasets. Inspired by random forests \cite{breiman2001random}, Liu \etal~\cite{liu2008isolation} employed different subsamples of training data to establish an ensemble of trees to isolate outliers on the basis of the path length from the root to the leaves. 

In recent years, with more attention focusing on outlier ensembles, several work discussed the theories and emphasized the crucial aspects of ensemble model construction. Aggarwal \cite{aggarwal2013outlier} and Zimek~\cite{Zimek13Ensemble} talked about algorithmic patterns, categorization, and the important building blocks of outlier ensembles such as model combination and diversity of base models. Zimek \etal~\cite{zimek2013subsampling} analytically and experimentally studied the subsampling technique and improved results through building an ensemble on top of several subsamples without mentioning the subsample selection. Aggarwal and Sathe\cite{aggarwal2015theoretical} deduced the bias-variance trade-off theory from classification to outlier detection, clarified some misconceptions about the existing subsampling methods, and proposed more effective subsampling and feature bagging approaches. On base detector combination, Rayana and Akoglu~\cite{rayana2015less} presented unsupervised strategies to select a subset of trusted detectors while omitting inaccurate ones in an unsupervised way.

Unlike  existing outlier ensembles that solely employ a sequential or parallel framework, our proposed method \care~ incorporates both of these building blocks to reduce both bias and variance. These two  phases respectively involve ($i$) successively eliminating outliers from the original dataset to build a better data model on which outlierness is estimated (sequentially), and ($ii$) combining the results from individual base detectors and across iterations (parallelly).

\section{Background and Preliminaries}
\label{sec:bp}

\vspace{-0.05in}
\subsection{Outlier Detection Problem}
\label{ssec:odp}
\vspace{-0.1in}
A popular characterization of an outlier is given by Hawkins as ``an observation which deviates so much from other observations as to arouse suspicion that it was generated by a different mechanism"~\cite{hawkins1980identification}. A common approach to outlier detection is to find unusual multi-dimensional points by quantifying a measure of normality relative to their neighboring points. Based on this notion, there exist two major varieties of outlier detectors: ($i$) distance, and ($ii$) density based. Specifically, distance based detectors find data points which are far from their nearest neighbors and density based detectors find the points which reside in a lower density region compared to their nearest neighbors. Formally, the problem can be stated as follows: \\
\noindent
{\bf Given} a multi-dimensional data $D$ with $n$ individual points in $d$ dimensions; \\
\noindent
{\bf Find} outliers which are far from the rest of the data (i.e., inliers) or reside in a lower density region.

\vspace{-0.05in}
\subsection{Bias-Variance Trade-off for Outlier Detection}
\label{ssec:bv_tradeoff}
\vspace{-0.1in}
The bias-variance trade-off is often explained in the context of supervised learning, e.g., classification, as quantification of bias-variance requires labeled data. Although outlier detection problems are solved using unsupervised approaches due to the lack of ground truth,
the bias-variance trade-off can be quantified similarly by treating the dependent variable (actual labels) as unobserved~\cite{aggarwal2015theoretical}.     

Unlike classification, most outlier detection algorithms output ``outlierness" scores for the data points. We can consider the outlier detection problem as a binary classification task having a majority class (inliers) and a minority class (outliers) by converting the outlierness scores to class labels. The points with scores above a threshold are considered as outliers with label 1 (label 0 for inliers below threshold). After converting the unsupervised outlier detection problem to a classification task with only unobserved labels, we can explain the bias-variance trade-off for outlier detection using ideas from classification. Specifically, the expected error of outlier detection can be split into two main components: reducible error and irreducible error (i.e., error due to noise). The reducible error can be minimized to maximize the accuracy of the detector. Furthermore, the reducible error can be decomposed into ($i$) error due to squared bias, and ($ii$) error due to variance. However, there is a trade-off while minimizing both these sources of errors.

 Bias of a detector is the amount by which the expected output of the detector differs from the actual label (unobserved) over the training data. While variance of a detector is the expected amount by which the output of a detector over one training set differs from the expected output of the detector over all the training sets. Simply put, the trade-off between bias and variance can be viewed as, ($i$) a detector with low bias is very flexible in fitting the data well and fits each training set differently with high variance, and ($ii$) an inflexible detector fits each training set almost similarly yielding high bias but low variance. 

\vspace{-0.05in}
\subsection{Motivation for Ensembles}
\label{ssec:motivEn}
\vspace{-0.1in}
Our goal in this work is to improve outlier detection by reducing both bias and variance, and in return decreasing the reducible error. It is evident from the classification ensemble literature that combining results from multiple base algorithms decreases the overall variance of the ensemble~\cite{brei96,lazarevic2005feature,aggarwal2015theoretical}, which is also true for outlier ensembles. On the other hand, this combination does not provide any evidence for reducing bias,  as controlled bias reduction is rather difficult due to lack of ground truth. However, there exist some successful heuristic approaches for reducing bias by removing outliers iteratively to build a more robust successive outlier detector. This iterative approach can be considered as a sequential ensemble. 
The basic idea here is that the outliers interfere with the creation of the normal data model, and the removal of points with high outlierness scores will be beneficial for the outlier model to produce an output close to the actual (unobserved) labels in expectation.

\section{Proposed Approach}
\label{sec:proposed}
\vspace{-0.05in}
\subsection{Overview}
\label{ssec:overview}
\vspace{-0.1in}
\care~ takes the $d$-dimensional data, a value for $k$ (nearest neighbor count), and a value for $MAXITER$ as input and outputs an outlierness score list $\textbf{fs}$ and a rank list $\textbf{r}$ (ranked based on most to least outlierness) of all the data points. In the experiments we use $k = 5$, which is compatible with the state-of-the-art methods~\cite{zimek2013subsampling,aggarwal2015theoretical}.
Moreover,  parameter $k$ in subsampling methods is scaled by the inverse of various subsample sizes,
as such large $k$ is not required (the smaller the subsample, the larger the relative neighborhood per point becomes for a fixed $k$). As for $MAXITER$, we set it to 15, a relatively small value.  We assume that our approach improves the base detectors over iterations, and the results are stabilized after only a few iterations and the algorithm stops following the stopping criterion.
\begin{algorithm}[!h]
	\caption{\care~Outlier Detection Ensemble}
	\label{alg:careAlgo}
	\begin{algorithmic}[1]
		\REQUIRE $d$-dimensional Data $D$, NN count $k = 5$, $MAXITER = 15$
		\ENSURE Score list ($\textbf{fs}$) and rank list ($\textbf{r}$) of points 
		\STATE $S = D$ (initially); $E = \emptyset$; $iter = 0$ 
		\WHILE{$iter \le MAXITER$}
		\STATE Obtain results from ($b$) feature-bagged base detectors ($D, S, k$) [Section~\ref{ssec:base}]
		\STATE Calculate pairwise agreement rates \textbf{$a_A$} for all base detector pairs in set $A$
		\STATE Estimate detector errors $\textbf{e}$ ($b \times 1$) based on \textbf{$a_A$} [Section~\ref{sssec:error}]
		\STATE Compute detector weights using estimated errors [Section~\ref{sssec:weight}]
		\STATE Compute pruned weighted outlierness scores of data points to get combined  scores $(\textbf{ws})$ [Section~\ref{sssec:weight}]
		\STATE $E = E \cup \textbf{ws}$
		\STATE $\textbf{fs} = average(E)$
		\STATE Generate new data sample $S$ from $D$ using \fvps~(w/o replacement) on $\textbf{fs}$ [Section~\ref{sssec:seqEn}]
		\IF{\textit{stopping condition} is TRUE} 
		\STATE $break$ [Section~\ref{sssec:SC}]
		\ENDIF
		\STATE $iter = iter + 1$
		\ENDWHILE
		\STATE $\textbf{r} = sort(\textbf{fs})$ (descending order)
	\end{algorithmic}
\end{algorithm}

The main steps of \care~are given in Algorithm \ref{alg:careAlgo}. 
Step 3 creates the feature-bagged outlier detectors as base detectors of the ensemble. For the first iteration the sample set $S$ contains the whole data $D$ as shown in step 1. For each base detector, $q \in [d/2, d-1]$ features are selected randomly to create the corresponding feature bag. We create $b$ ($=100$) feature-bagged base detectors. Motivated by Platanios \etal~\cite{platanios2014estimating}, step 4 calculates the pairwise agreements \textbf{$a_A$} for all possible pairs of base detectors and step 5 estimates the error rates of the individual base detectors in an unsupervised way using \textbf{$a_A$}. Step 6 calculates weights for the base detectors using their corresponding error rates. Step 7 combines the outlierness scores from the different base detectors with weighted average combination to get final outlierness scores $\textbf{ws}$. Step 8 stores the outlierness scores in $E$ at each iteration and step 9 calculates the final outlierness scores $\textbf{fs}$ by averaging the results of all previous iterations as well as the current iteration. Based on $\textbf{fs}$, step 10 generates the new data sample $S$ (where, $|S| < |D|$) using the \fvps~ approach w/o replacement (see Section~\ref{sssec:seqEn})  and step 11 generates the ranked list $\textbf{r}$ of instances from most to least outlierness. We repeat steps 3-16 until the stopping condition at step 12 is met or upto the given maximum iteration $MAXITER$.

Unlike existing ensemble techniques, \care~incorporates a two-phase aggregation approach in each iteration; first, it combines the results from the individual base detectors (parallel) and second, it cumulatively aggregates the results from multiple iterations (sequential).

Next we describe the main components of our proposed \care~in detail. In particular, we describe the base detectors in Section \ref{ssec:base} and consensus approaches in Section \ref{ssec:consensus}. 

\vspace{-0.05in}
\subsection{Base Detectors}
\label{ssec:base}
\vspace{-0.1in}
There exist various approaches for outlier detection based on different aspects of outliers, designed for distinct applications to detect domain-specific outliers. In our work, we are interested in \textit{unsupervised} outlier detection approaches that assign outlierness scores to the individual instances in the data, as such, allow ranking of instances based on outlierness.

\vspace{0.025in}
\subsubsection{kNN based Outlier Detectors}
\label{sssec:od}
\vspace{-0.1in}
There are a number of well-known unsupervised approaches, e.g., ``distance based'' and ``density based'' methods for outlier detection. Distance based methods~\cite{knorr1997unified,zhang2009new} and their variants try to find the \textit{global} outliers far from the rest of the data based on $k$ nearest neighbor ($kNN$) distances of the data points. On the other hand, density based methods~\cite{breunig2000lof,papadimitriou2003loci,kriegel2009loop} and their variants try to find the \textit{local} outliers which are located in a lower density region compared to their $k$ nearest neighbors. 

In this work, we create two versions of \care: ($1$) using the distance based approach \avgknn~(average $k$ nearest neighbor distance of individual data point is used as outlierness score), and ($2$) using the most popular density-based approach \lof~\cite{breunig2000lof}. We note that \care~is flexible to accommodate any other nearest neighbor based outlier detection algorithm as well.

\subsubsection{Feature Bagging}
\label{sssec:fb}
\vspace{-0.1in}
Feature bagging is commonly used in classification ensembles for dimensionality reduction as well as for variance reduction. Like classification ensembles, feature bagging can also be incorporated in outlier ensembles in order to explore multiple subspaces of the data to induce diverse base detectors for high-dimensional outlier detection. As such, in this work we incorporate feature bagging to create multiple base detectors and combine their results to detect outliers with a goal to improve the detection performance by reducing variance. Given a $d$-dimensional dataset $D$, for each base detector (either \lof~or \avgknn), we randomly select $q \in [d/2, d-1]$ features  to create $b$ ($=100$) different feature-bagged base detectors.

\subsection{Consensus Approaches}
\label{ssec:consensus}
\vspace{-0.1in}
Unlike classification, building an effective ensemble for outlier detection is a challenging task due to the lack of ground truth, which makes it difficult to measure the detector accuracy and combine the results from accurate detectors. Most of the existing approaches either combine outcomes of all the base detectors~\cite{lazarevic2005feature,RayanaAkoglu14} (hurting the ensemble in presence of poor detectors), or selectively incorporate accurate base detectors in an unsupervised fashion discarding the poor ones~\cite{Zimek13Ensemble,rayana2015less}. However, the definition of a poor detector varies across different application domains, as some selective approaches are useful for certain applications but not as useful for others. Therefore, in this work we go beyond binary selection and estimate weights for individual base detectors to aggregate their results with a weighted combination. Furthermore, we cumulatively combine the weighted aggregation results for multiple iterations until a stopping condition is satisfied. In the following two sections, we describe the error estimation and weighted aggregation of the base detectors. In Section~\ref{sssec:seqEn} the sequential aggregation approach is described and in Section~\ref{sssec:SC} we introduce a stopping condition for our iterative \care~approach.

\subsubsection{Error Estimation}
\label{sssec:error}
~Platanios \etal~\cite{platanios2014estimating} proposed an \textit{unsupervised} approach called Agreement Rates (AR) to estimate errors of multiple classifiers. Motivated by~\cite{platanios2014estimating}, we adapt the unsupervised error estimation of the individual outlier detectors in our work. This estimation is based on the agreement rates for all possible pairs of base detectors in $A$. Outlier detection can be considered as a binary classification problem with a majority class (inliers $= 0$) and a minority class (outliers $= 1$). However, most  existing outlier detection algorithms provide outlierness scores for the data points, and not $\{0,1\}$ labels for them. In order to adapt the AR approach, we calculate the agreement rates for all possible pairs of detectors in $A$, for which $\{0,1\}$ labels are needed for the data points. We use Cantelli's inequality~\cite{grimmet2001} to estimate a threshold $th_i ~(i = 1 \ldots b)$ with confidence level at $20\%$ to find a cutoff point between inliers ($=0$) and outliers ($=1$) for each base detector to get a binary list of class labels. 

After estimating the class labels, we calculate the agreement rates. As inliers are the majority class, if we take into account all the data points in calculating the agreement rates, it is likely that most values would be large as most detectors often agree on a large number of inliers. Our main goal is to find agreement based on the outliers detected by the base detectors, and ignore a large number of inliers. Therefore, we take the union of all outliers ($=1$) across different base detectors to obtain $U$.  Set $U$ contains the important data points (detected as outliers), which we use to calculate the agreement rates for the detector pairs in $A$.

In the following sections we denote the base detectors as $f_i \in F ~ (i = 1 \ldots b, |F|=b)$, input data as $D$, and class labels as $Y$. The error event $E_A$ of a set of detectors in $A$ is defined as an event when all the detectors make an error: \\
\vspace{-0.05in}
\begin{equation}
	E_A = \bigcap_{i \in A}[f_i(D) \neq Y] ~,
\end{equation}
where $\bigcap$ denotes set intersection. The error rate of a set of detectors in $A$ is then defined as the probability that all detectors in $A$ make an error together and is denoted as \\
\vspace{-0.05in}
\begin{equation}
e_A = \mathbb{P}(E_A) ~.
\end{equation}
The agreement rate of two detectors  is the probability that both make an error or neither makes an error. As such, the pairwise agreement rate equation in terms of error rates for the sets in $A: |A| = 2$ can be written as  \\
\begin{equation}
\label{agree}
\begin{split}
a_{\{i,j\}} & = \mathbb{P}( E_{\{i\}} \cap E_{\{j\}} ) + \mathbb{P}( \bar{E}_{\{i\}} \cap \bar{E}_{\{j\}} )  \\
& = 1 - e_{\{i\}} - e_{\{j\}} + 2 e_{\{i,j\}} ~,~~ \forall \{i,j\} \in A : i \neq j ~,
\end{split}
\end{equation}
where $\bar{\cdot}$ denotes the set complement.
On the other hand, the agreement rates for the set of detectors in $A: |A| = 2$ can be directly calculated from the detector output and set $U$ (defined earlier) as follows: \\
\vspace{-0.05in}
\begin{equation}
a_{A} = \frac{1}{|U|}\sum_{u = 1}^{|U|}\mathbb{I}\{f_i(D_u) = f_j(D_u)\},\forall\{i,j\} \in A : i \neq j ~.
\end{equation}

Provided that one can easily compute the pairwise agreement rates $a_{\{i,j\}}$'s, which can be written in terms of the (unknown) individual and pairwise error rates of the detectors, we can 
cast the error rate estimation as a constrained optimization problem where the agreement equations in \eqref{agree}  form constraints that must be satisfied as follows:
\begin{equation}
\label{optim}
\begin{split}
\mathbf{min.} ~~ & \sum_{\hat{A}: |\hat{A}| \le 2} e_{\hat{A}}^{2} + {\epsilon_{\hat{A}}} \\
{\mathbf{s.t.}}~~ & a_{A} = 1 - e_{\{i\}} - e_{\{j\}} + 2 e_{\{i,j\}} ~,~ \forall \{i,j\} \in A \\
			& 0 \le e_{\hat{A}} < 0.5 + \epsilon_{\hat{A}}  ~,~~\\
			&  0 \le \epsilon_{\hat{A}}\\
 \end{split}
\end{equation}
where $\hat{A}$ contains individual as well as pairs of detectors (i.e., $\hat{A} = F \cup A$) and  $\epsilon_{\hat{A}}$'s denote the slack variables.

\begin{figure}[h]
	\vspace{-0.15in}
	\centering
	\begin{tabular}{cc}
		\includegraphics[width=0.45\linewidth,height=1.15in]{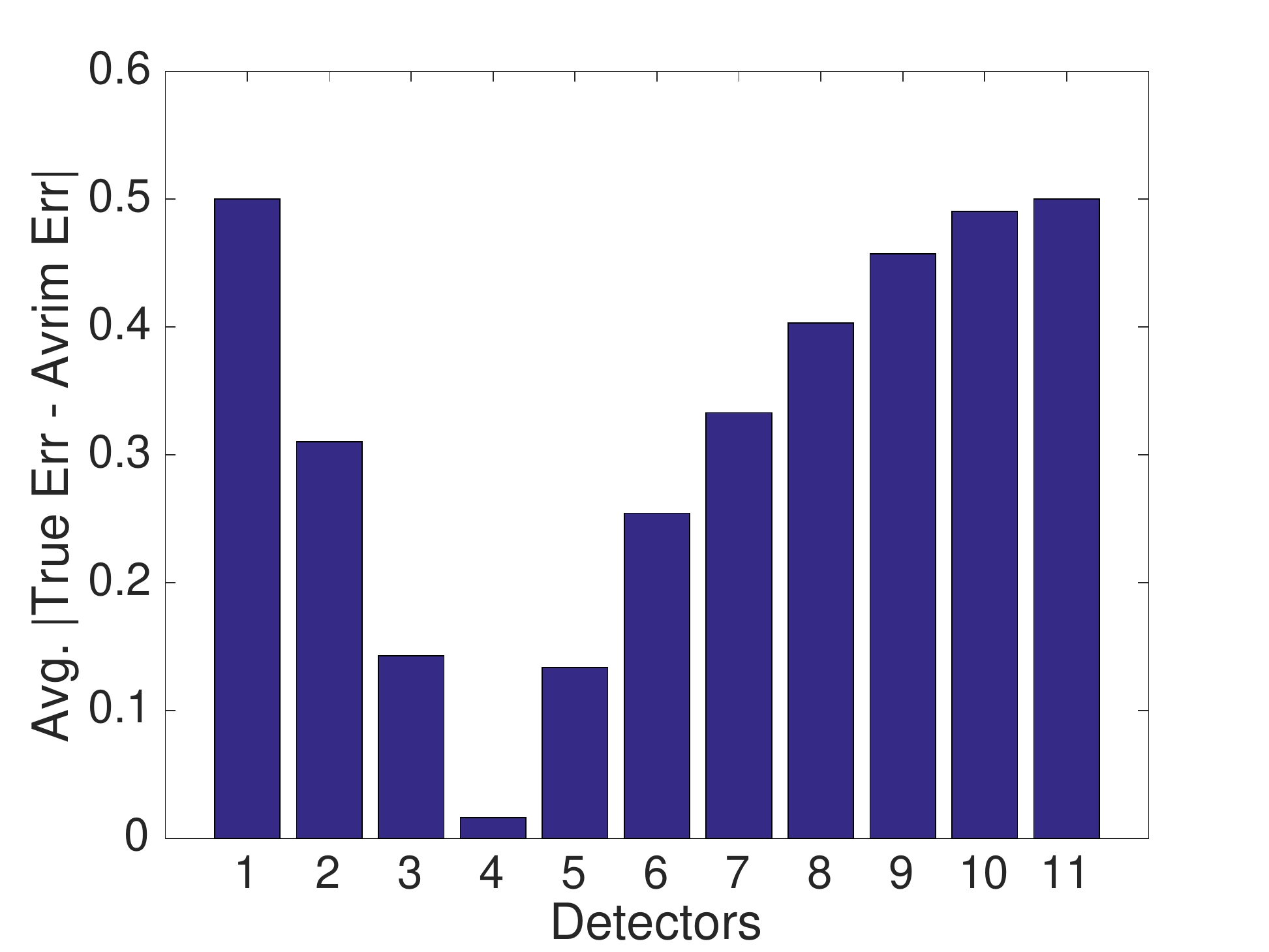}&
		\includegraphics[width=0.45\linewidth,height=1.15in]{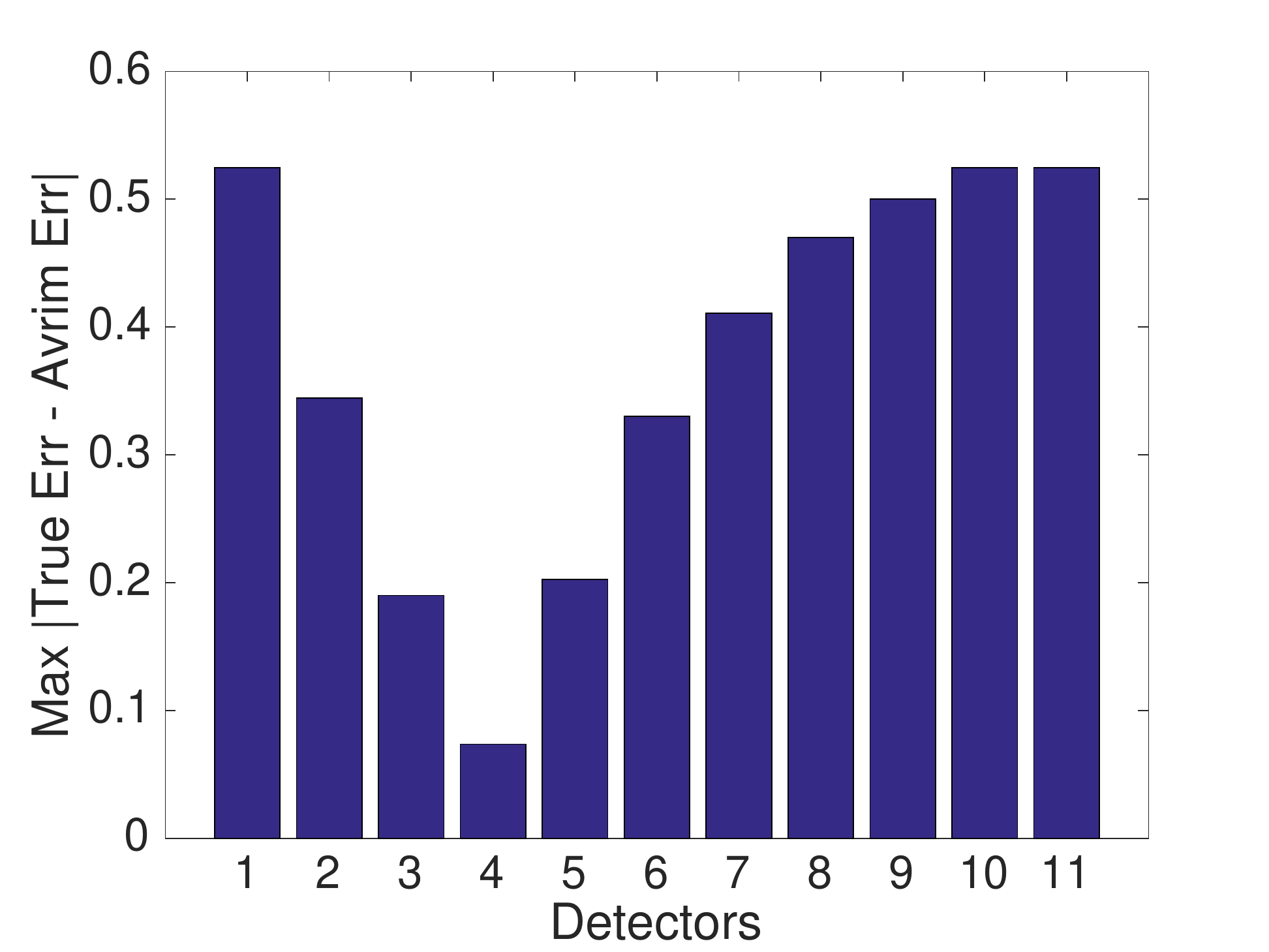}\\
		\multicolumn{2}{c} \;(a) $e  = [0.0,0.1,0.2,0.3,0.4,0.5,0.6,0.7,0.8,0.9,1.0]$ \\
		\includegraphics[width=0.45\linewidth,height=1.15in]{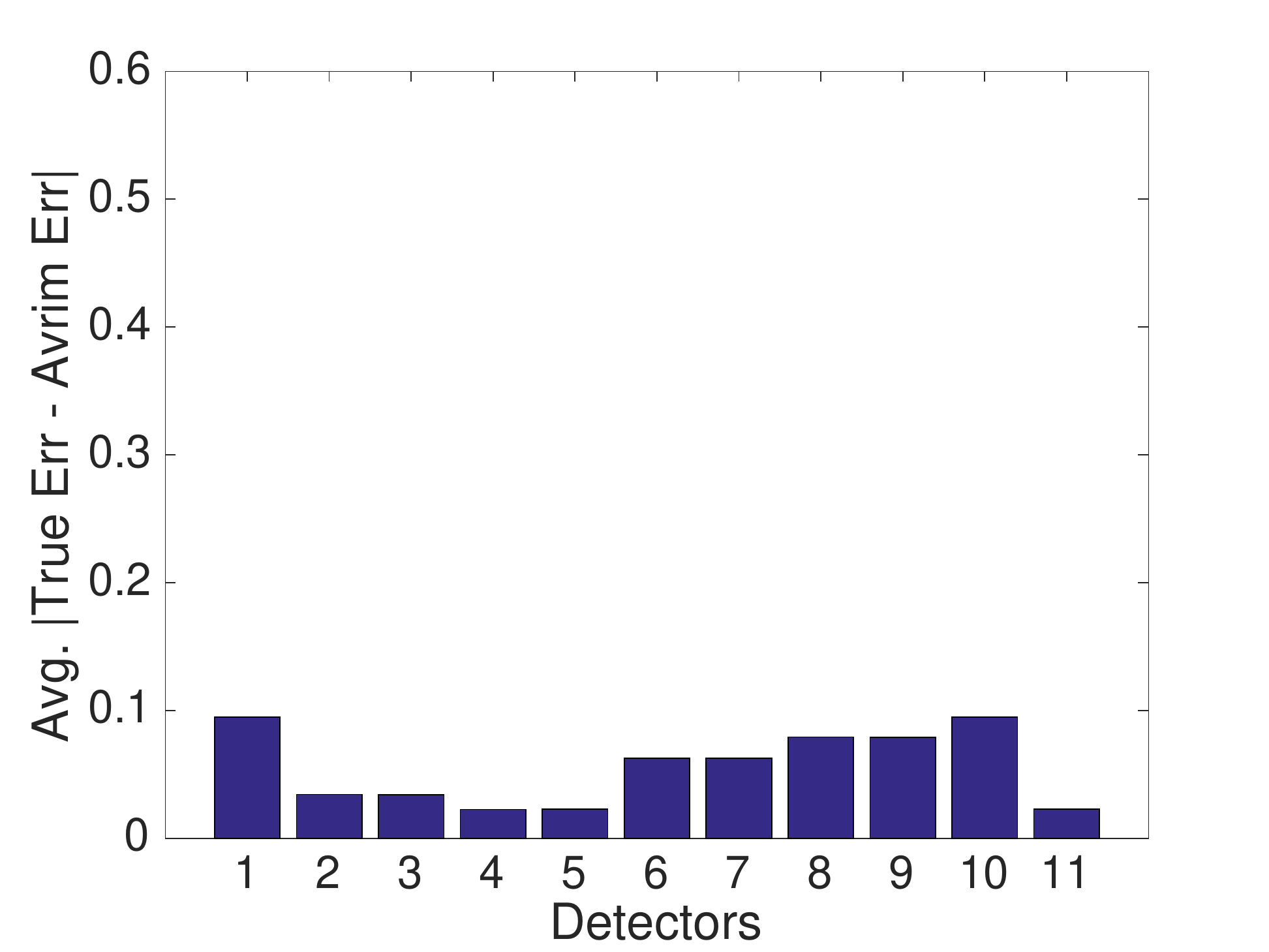}&
		\includegraphics[width=0.45\linewidth,height=1.15in]{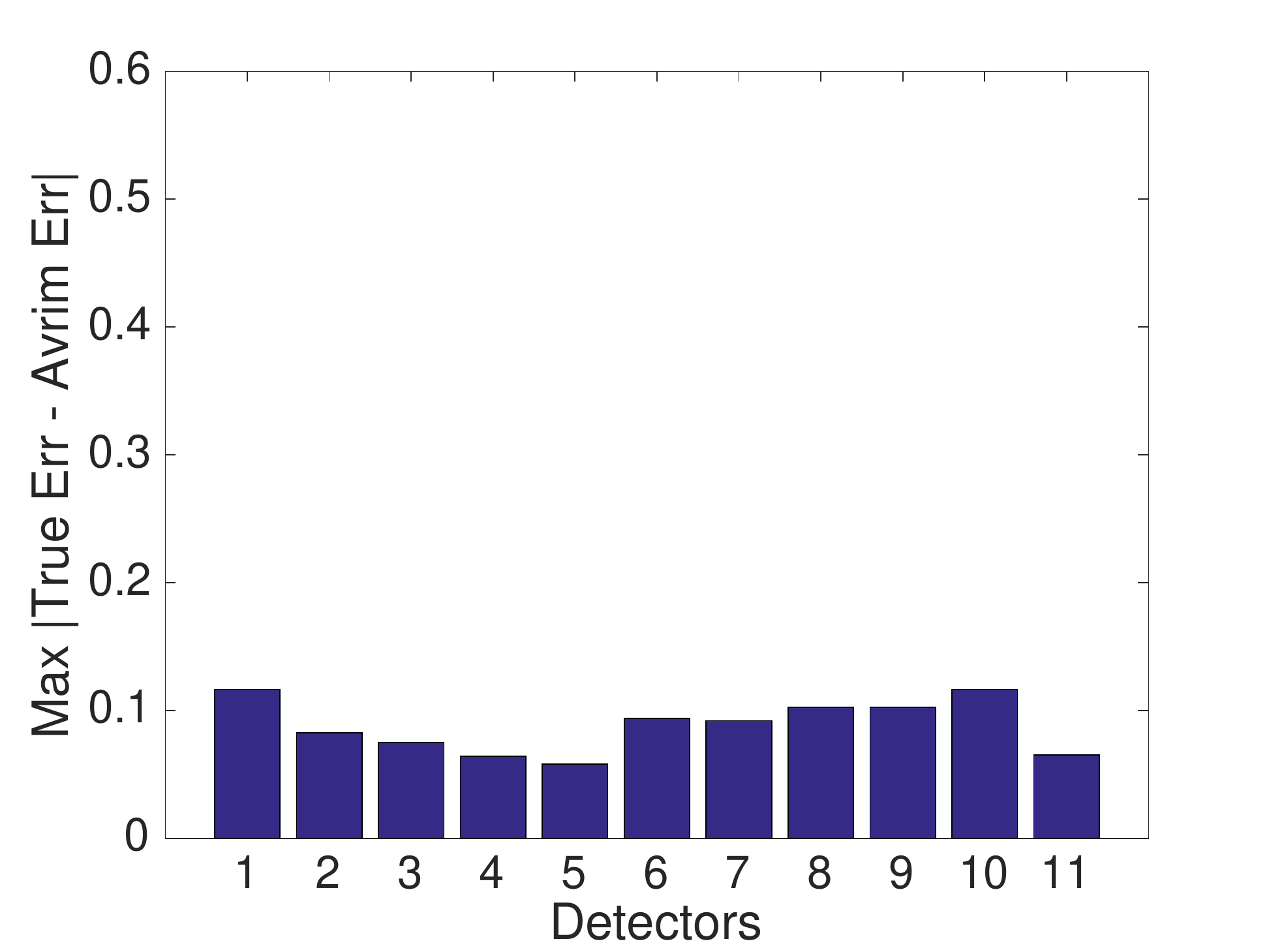}\\
		\multicolumn{2}{c} \;(b) $e = [0.0,0.0,0.1,0.1,0.2,0.2,0.2,0.3,0.3,0.4,0.4]$ \\
		\includegraphics[width=0.45\linewidth,height=1.15in]{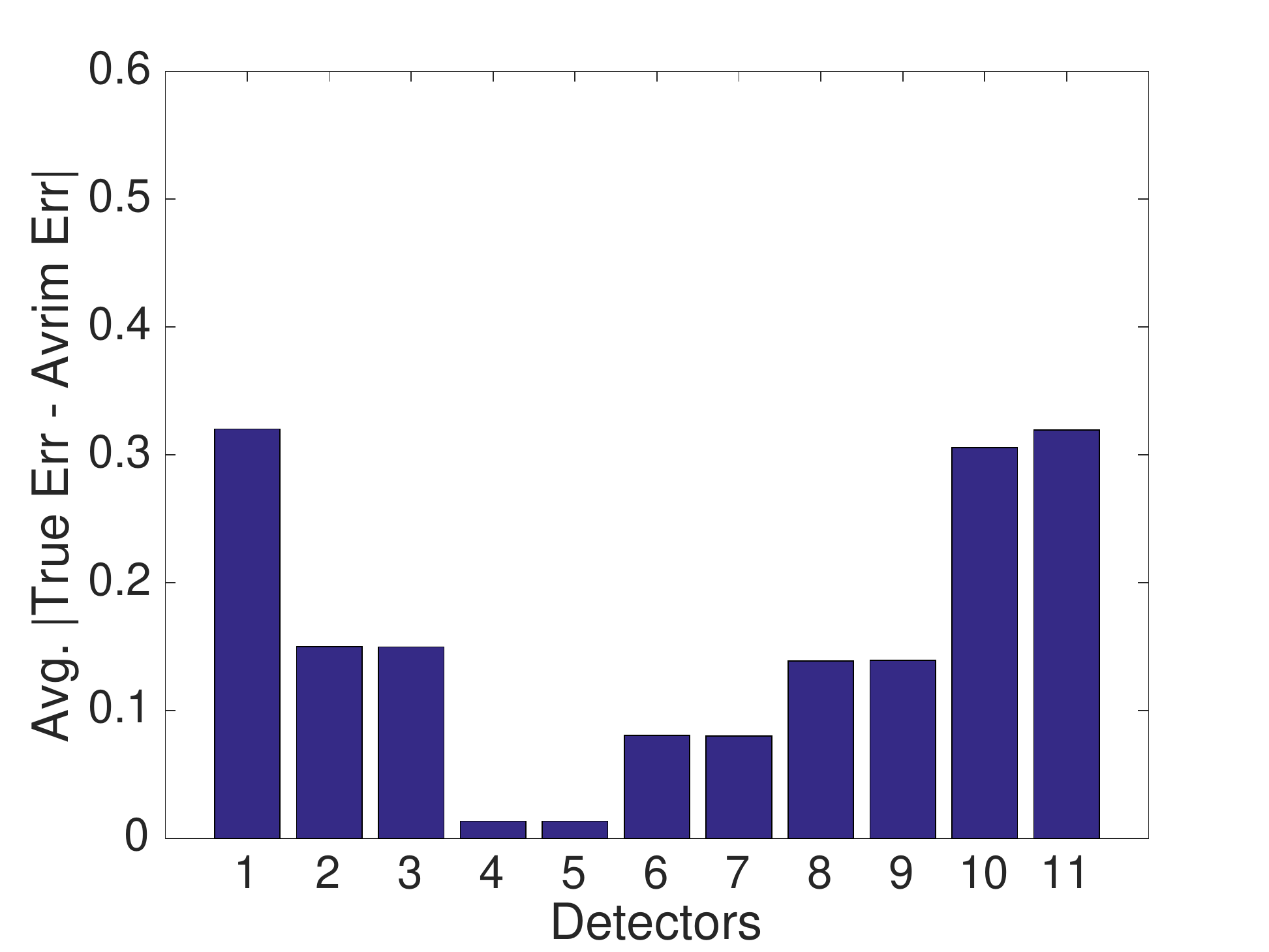}&
		\includegraphics[width=0.45\linewidth,height=1.15in]{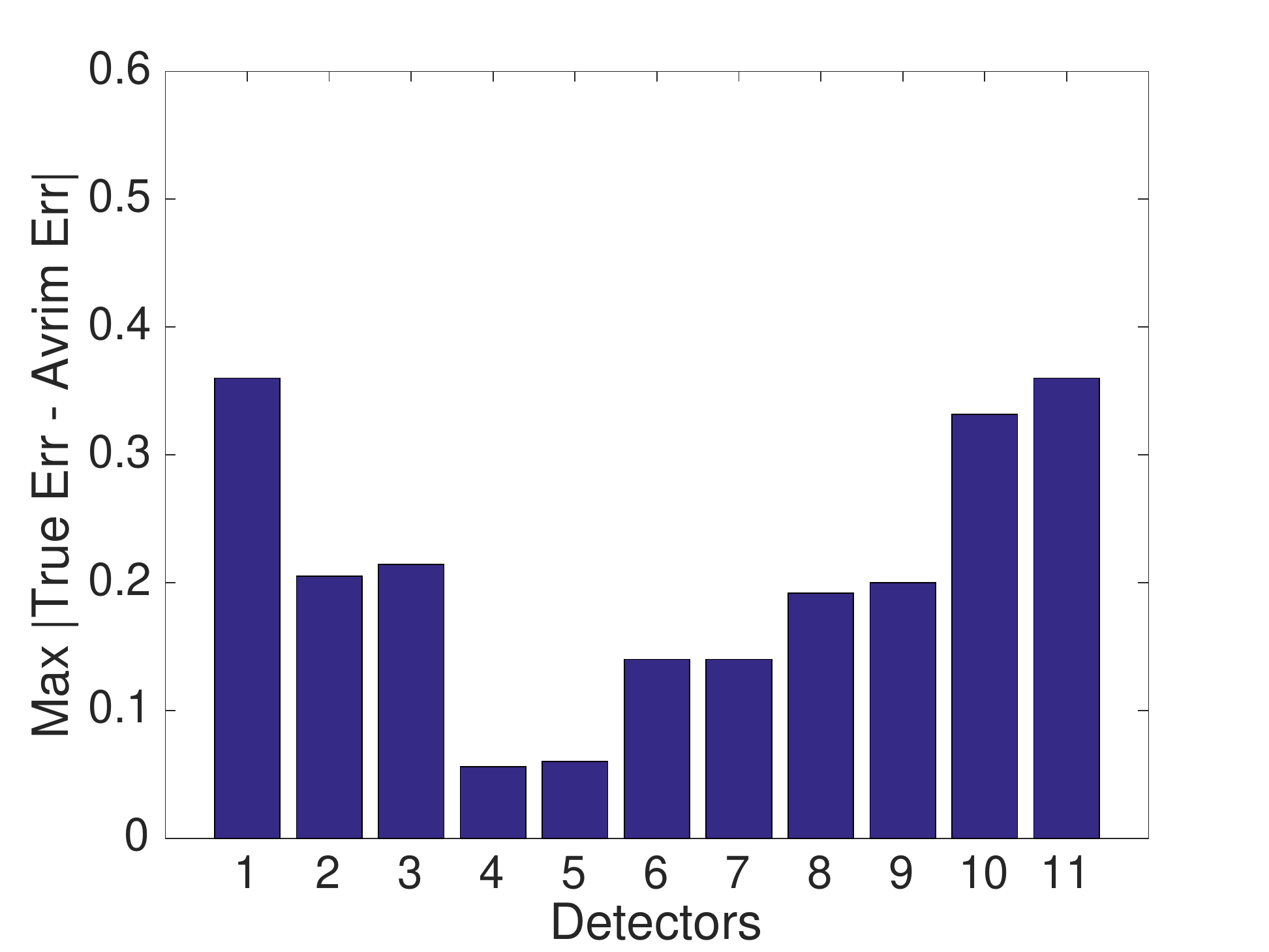}\\
		\multicolumn{2}{c} \;(c) $e = [0.0,0.1,0.1,0.2,0.2,0.3,0.3,0.4,0.4,0.8,0.9]$ \\
		\end{tabular}
	\vspace{-0.05in}
	\caption{Average (left) and Maximum (right) difference (across datasets) between the true error and the estimated error of different base detectors ($e$ represents true error rates of base detectors). Notice that the differences are high with the presence of many bad detectors, but low otherwise.
		\label{fig:errorrate}}
\end{figure}

In their AR approach, Platanios \etal~ assume that the error rates should be strictly $< 0.5$. Different from theirs, we allow the error rates to be above $0.5$, for which we introduce a slack variable $\epsilon_{\hat{A}} \ge 0$ in  constraints $0 \le e_{\hat{A}} \le 0.5 + \epsilon_{\hat{A}}$. In real-world settings, it is possible to have poor base detectors having large errors (i.e., worse than random). Then the question becomes whether the presence of poor detectors (with error $\ge 0.5$)  hampers the overall estimation of the errors. To answer this question we have designed experiments with synthetic datasets mimicking the real datasets having 1000 data points in total, where 10\% of them are outliers and 11 base detectors with different true error rates. 
We generate multiple ($=100$) snapshots of the synthetic datasets randomly, to analyze results. Figure~\ref{fig:errorrate} shows the average and maximum difference between the true error and estimated error of different base detectors. In Figure~\ref{fig:errorrate} (a), 6/11 detectors have true error $\ge 0.5$, as a result the average and maximum differences are larger as compared to Figure~\ref{fig:errorrate} (b) and (c), where in (b) none and in (c) only 2/11 detectors have errors $\ge 0.5$. We conclude that if all the detectors are good (better than random) or only a few are bad, the optimization in \eqref{optim}  estimates meaningful error rates.

Although the above constrained optimization approach estimates error rates of individual as well as of all possible pairs of base detectors, we only utilize the error rates of the individual detectors to calculate their corresponding weights for aggregation, which we describe next.

\subsubsection{Weighted Aggregation}
\label{sssec:weight}
Most commonly used aggregation functions in outlier ensembles are \textit{average} and \textit{maximum}. In most cases, average is preferred over maximum as the latter overestimates the absolute scores. On the other hand, averaging might dilute the final scores with the presence of poor detectors. In \care, we propose to use \textit{weighted aggregation} to improve the ensemble. 

\begin{figure}[h]
	\vspace{-0.2in}
	\centering
	\begin{tabular}{cc}
		\hspace{-0.1in}\includegraphics[width=0.52\linewidth,height=1.5in]{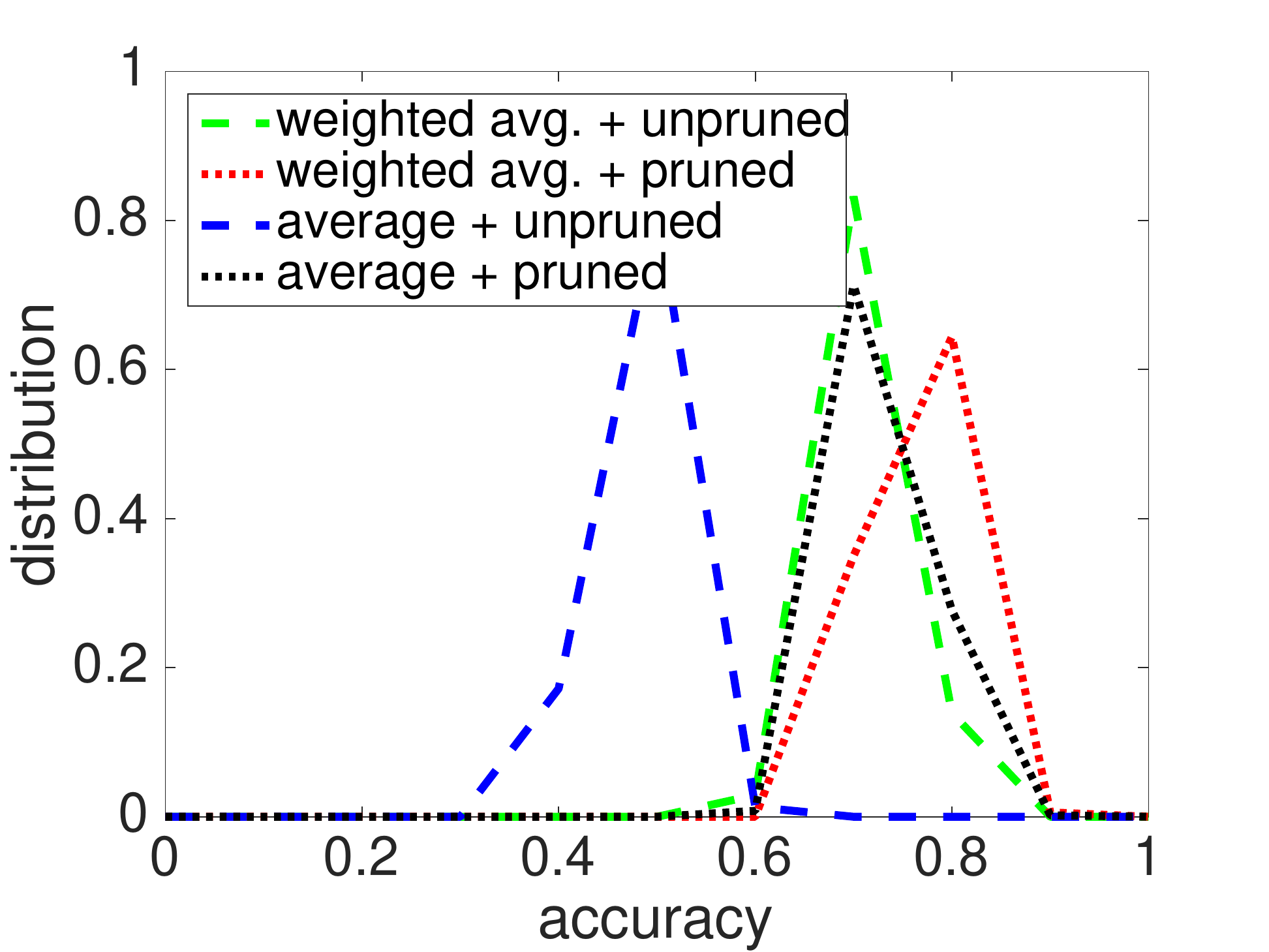}&
		\hspace{-0.2in}\includegraphics[width=0.52\linewidth,height=1.5in]{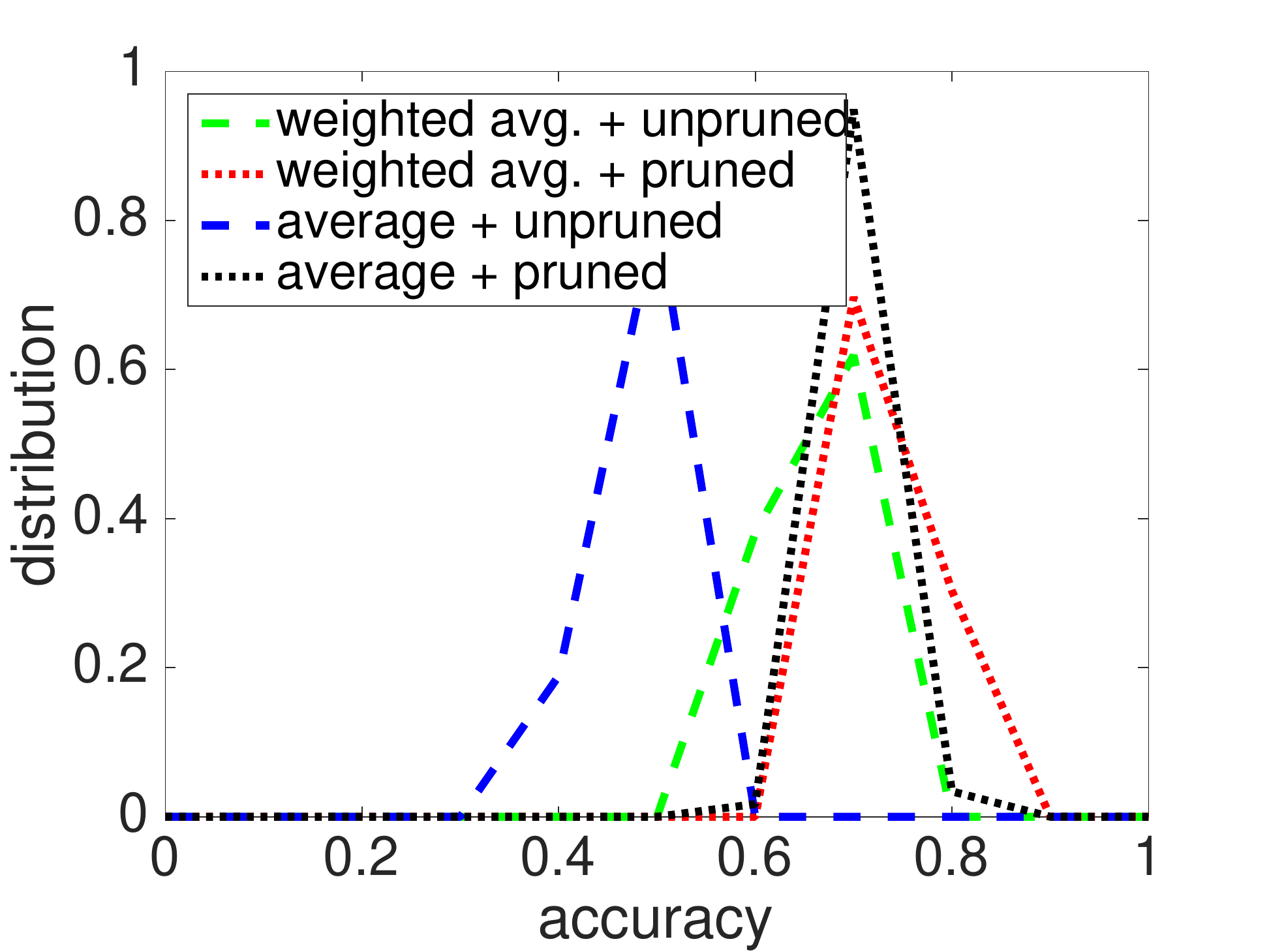}\\
		5\% outliers & 10\% outliers \\
		\end{tabular}
	\vspace{-0.1in}
	\caption{Distribution of accuracies with different consensus approaches. Notice the distribution with pruned weighted aggregation (red curve) is skewed towards higher accuracies. 
		\label{fig:wgp}}
	\vspace{-0.05in}
\end{figure}

We calculate the weights of the base detectors from their estimated error rates as described in the previous section, such that the weights are positive and inversely proportional to the corresponding errors. Inspired by AdaBoost~\cite{freund1997decision}, we employ the error rates of individual detectors to calculate their corresponding weights using the following equation: \\
\vspace{-0.05in}
\begin{equation}
w_{i} = \frac{1}{2}\log{\Big(\frac{2}{e_i} - 1\Big)} ~, ~ i = 1 \ldots b
\end{equation}
where $w_i \ge 0$ is the weight of detector $i$ with estimated error $e_i \in [0, 1]$, for $i = 1 \ldots b$. Moreover, as we assume that in real-world settings the base detector pool will have poor (i.e., worse than random) detectors, we also incorporate a pruning strategy where we discard the detectors with error $e_i \ge 0.5$. To support the weighted aggregation strategy with pruning, we provide experimental results with synthetic datasets to compare average vs. weighted aggregation as well as pruned vs. un-pruned selection. In Figure~\ref{fig:wgp}, we show the distributions of the final ensemble accuracies with different consensus approaches across 1000 samples of a synthetic dataset. Each synthetic data has 1000 points and 4 detectors with true errors $e = [0.2,0.4,0.6,0.8]$ having 5\% (left) and 10\% (right) of outliers respectively. Here, we calculate the weights and prune the detectors using these estimated errors. We can see from both figures that weighted consensus is better than averaging and pruning is better than un-pruned aggregation as the red curve is more skewed towards the higher accuracy values than the others.

After pruning $p$ detectors with error $e_i \ge 0.5$, we combine the outlierness scores from the base detectors using weighted aggregation. In order to do weighted aggregation, we need to unify the outlierness scores, as different base detectors employ different feature sets, hence provide scores with varying range and scale. To standardize, we use Gaussian Scaling~\cite{Zimek11Uni} to convert the outlierness scores of \avgknn~or \lof~into probability estimates $Pr_i ~ (i = 1 \ldots b-p) \in [0, 1]$. We calculate the final outlierness score $ws({x})$ of a data point $x$ using the weighted average of the probability estimates as follows: \\
\vspace{-0.05in}
\begin{equation}
ws({x}) = \frac{\sum_{i=1}^{b-p}w_{i} \times Pr_{i}(x)}{\sum_{i=1}^{b-p}w_{i}} 
\end{equation}  
Above, $\sum_{i=1}^{b-p}w_{i}$ is used to normalize the outlierness scores. The final  scores can be used to sort the data points from most to least outlierness to produce a ranked list. 

Thus far, we described steps 3--7 of Algorithm \ref{alg:careAlgo}.
Next we describe the iterative nature of our sequential ensemble.

\subsubsection{Sequential Ensemble}
\label{sssec:seqEn}
With the weighted aggregation combining multiple feature-bagged base detectors we aim to reduce variance, but our additional goal is to reduce bias. In outlier detection, it is hard to reduce bias in a controlled way, but there exist some successful heuristics to reduce bias. One commonly used  approach is to remove outliers in successive iterations~\cite{journals/sigkdd/Aggarwal12} in order to build more robust outlier models iteratively. This is a type of sequential ensemble. The basic idea is that the outliers interfere with the creation of a model of normal data, and the removal of points with high outlier scores is beneficial for the model in the following iteration.  

As such, we adopt a sequential ensemble approach in \care~ where we use the result from the previous iteration to improve the next. In particular, we select a subsample $S$ from the original data $D$ (where $|S| < |D|$) to use it as a \textit{new data model based on which we calculate the outlierness scores} for all the data points in $D$. For example, when we need the average $k$NN distance of a data point $x \in D$, we calculate the distance to its $k$-nearest neighbors $N_i \in S$. The goal is to construct $S$ that includes as few of the true outliers as possible, such that it serves as a more reliable data model. To do so, we design a sampling approach which we call Filtered Variable Probability Sampling (\fvps).  Following are the steps of the \fvps: \\
\vspace{-0.175in}
\begin{itemize}
\setlength{\itemsep}{-0.75\itemsep}
\item Discard top $T$ outliers detected in previous step from $D$, where $T$ is the number of outliers selected using Cantelli's inequality~\cite{grimmet2001} on final outlierness scores $\textbf{fs}$ (threshold is selected at $20\%$ confidence level to find the cutoff point between outliers and inliers).
\item Select $l$ uniformly at random between $\min\{1-\frac{T}{n},\frac{50}{n}\}$ and $\max\{1-\frac{T}{n},\frac{1000}{n}\}$, where $n$ is the number of points in the original dataset.
\item Build sub-sample $S$ (where $|S| = l\times(n-T)$) by sampling from $D'$ (outliers-discarded)  based on the probability of the points being normal (i.e., $(1 - \textbf{fs})$).
\end{itemize}

In step 1 of \fvps, we obtain $D'$ by filtering the outliers detected in the previous step to reduce bias and improve the outlier ensemble iteratively. Here, we choose confidence level $20\%$ to get a larger $T$ in order to remove as many outliers as possible. Even though this step might remove some inliers, those should not effect the model as they would have lower probability of being normal points to be removed in the first place. Inspired by Aggarwal and Sathe~\cite{aggarwal2015theoretical}, we adopt variable sampling to select a sample size in step 2. The variable sampling approach has an effect over the parameter choice of the outlier detectors (i.e., $k$). Varying the subsample size at fixed $k$ effectively varies the percentile value of $k$ in the subsample for different iterations. For some datasets smaller value of $k$ is better, for others larger is better. However, there is no known suitable approach to estimate the correct value of $k$ for a dataset. Therefore, in \care~we select a small value of $k$ (e.g., $5$) and employ variable sampling to incorporate the illusion of using different $k$ values in different iterations, which introduces diverse detectors iteratively. 
After deciding the sample size in step 2, we use probability sampling to create the data model $S$ in step 3. Here, we choose a point from $D'$ to include  in $S$ based on its probability of being normal. As a result, we expect to have less  interference from the outliers in  $S$ as it is mostly built with normal points.

\fvps~introduces diverse detectors based on different $S$ in each iteration, hence, we aggregate (e.g. cumulative average) the outlierness scores \textbf{ws} over the iterations to compute final scores \textbf{fs} to further reduce variance and improve the sequential ensemble (step 9 in Alg. \ref{alg:careAlgo}). 
Note that \textbf{fs} is also what \fvps~uses for discarding outliers and sampling set $S$.

\vspace{-0.05in}
\subsubsection{Stopping Criterion}
\label{sssec:SC}

We need a proper stopping criterion for the sequential ensemble approach to decide where the iteration should stop and return the final result. As the whole framework is unsupervised, there is no way to use intermediate evaluation to find a stopping point. In \care, we utilize the pairwise agreement rates \textbf{$a_A$} between all possible pairs of base detectors to find the stopping point. Experiments reveal a useful strategy: if the distribution of \textbf{$a_A$}'s is skewed towards higher agreement rates, then the error estimates of the base detectors tend to be more accurate. Intuition is, it is unlikely that most pairs would have high agreement and yet agree on the wrong labels. 
Therefore, we propose to track the agreement rates and their distribution to decide when to stop iterating \care.

Specifically, we use the area under the curve ($auc$) of the {complementary cumulative distribution function ($ccdf$)} of \textbf{$a_A$}'s as the quantitative measure to decide the stopping point. The $auc$ of $ccdf$ is large if the distribution of \textbf{$a_A$}'s is skewed towards higher agreement rates and vice versa. We assume that as \care~sequentially progresses over iterations, the base detectors improve, and hence the $auc$ of $ccdf$ for pairwise agreement rates gets larger. However, if at any iteration $t+1, t \in [0, MAXITER]$, the $auc(t+1)$ falls below the average by more than the standard deviation of $auc(0,\ldots, t+1)$, the sequential ensemble stops and returns the result at iteration $t$ or otherwise iterates until $MAXITER$ and returns the final result.

\vspace{-0.05in}
\section{Reducing Bias and Variance with \lcare}
\label{sec:bv}

\vspace{-0.05in}

\begin{figure}[t]
	\centering
	\begin{tabular}{cc}
		\hspace{-0.1in}\includegraphics[width=0.53\linewidth,height=1.5in]{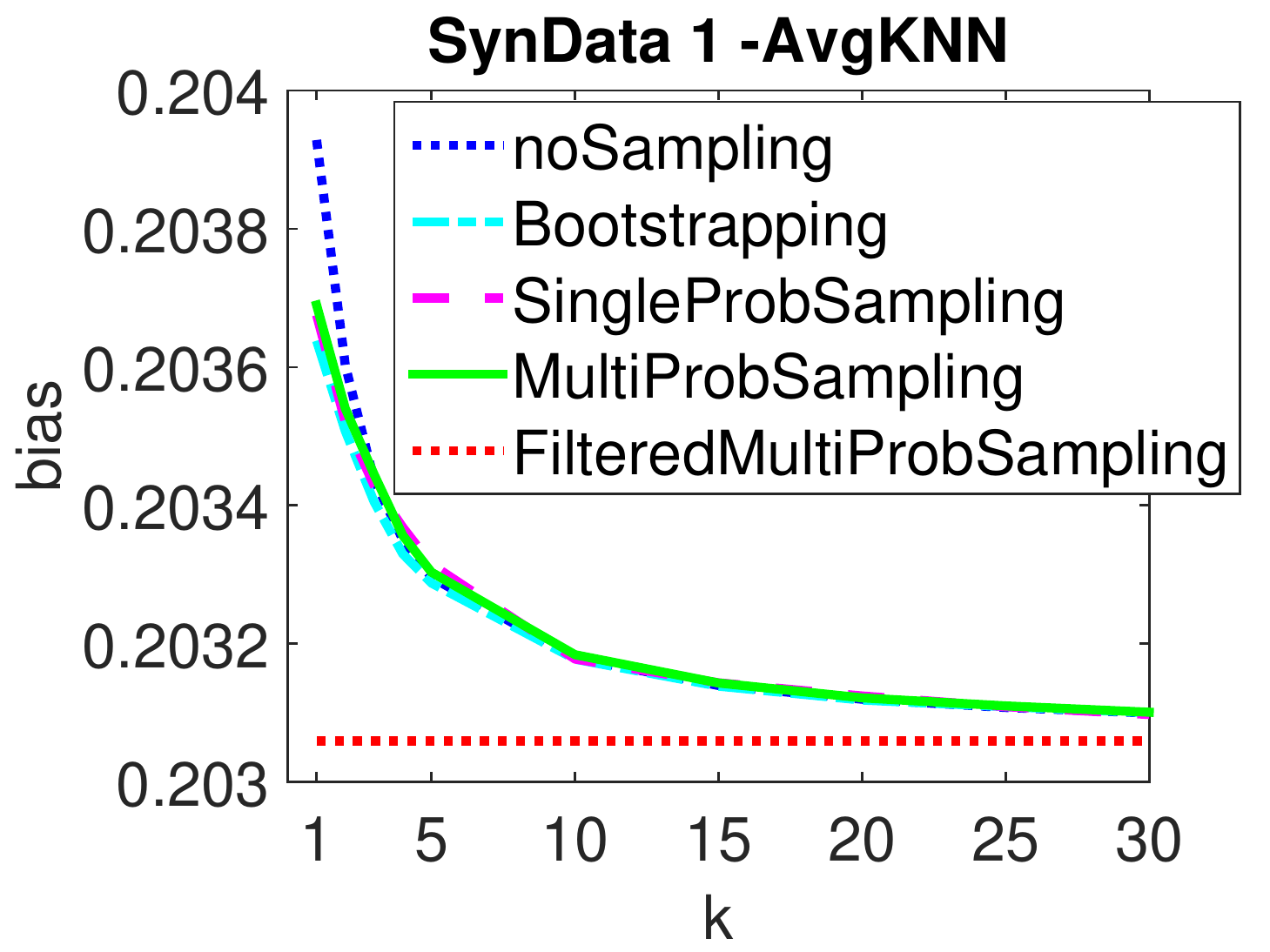}&
		\hspace{-0.2in}\includegraphics[width=0.51\linewidth,height=1.5in]{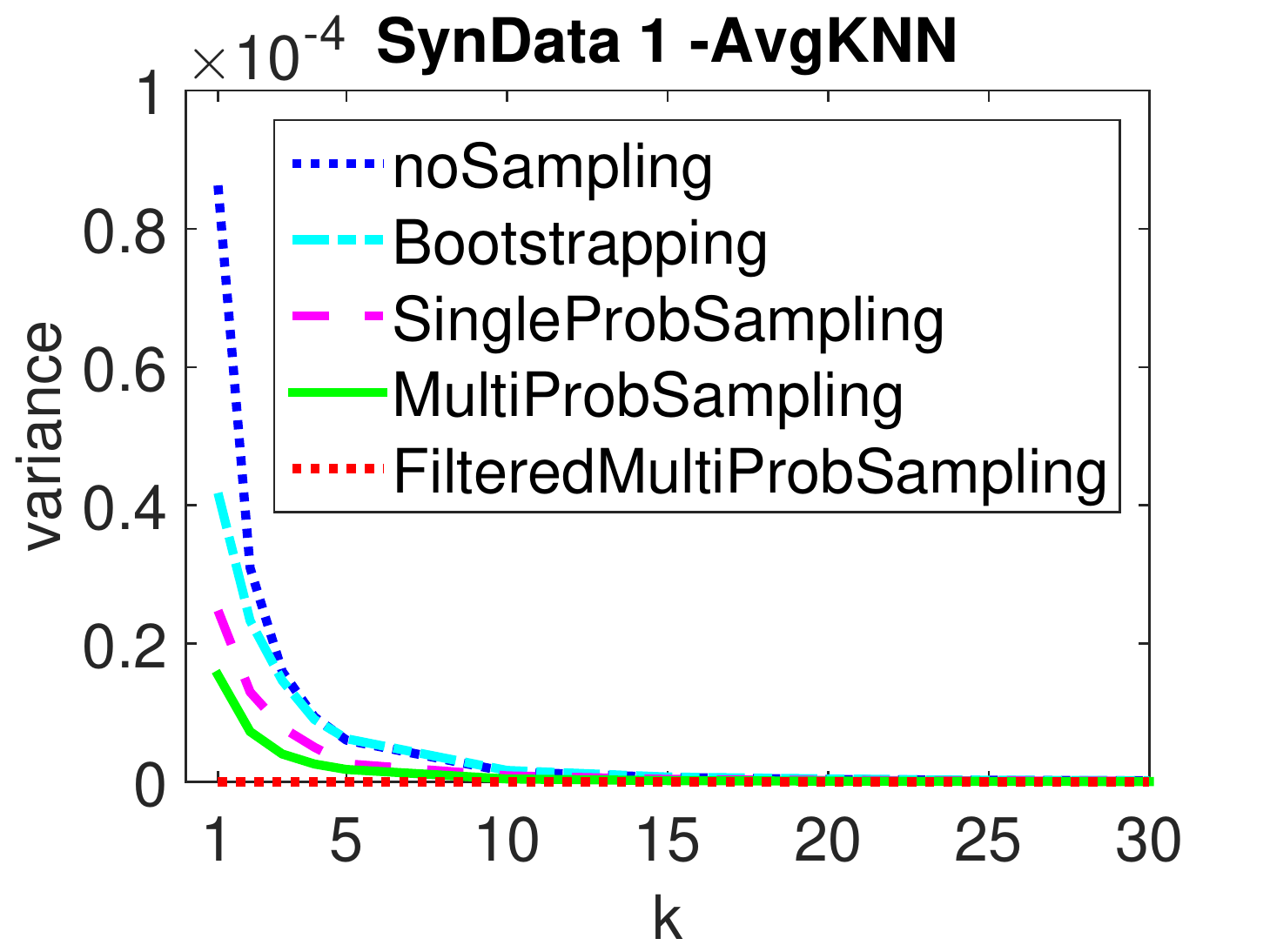}\\
		(a) & (b) \\
		\hspace{-0.1in}\includegraphics[width=0.53\linewidth,height=1.5in]{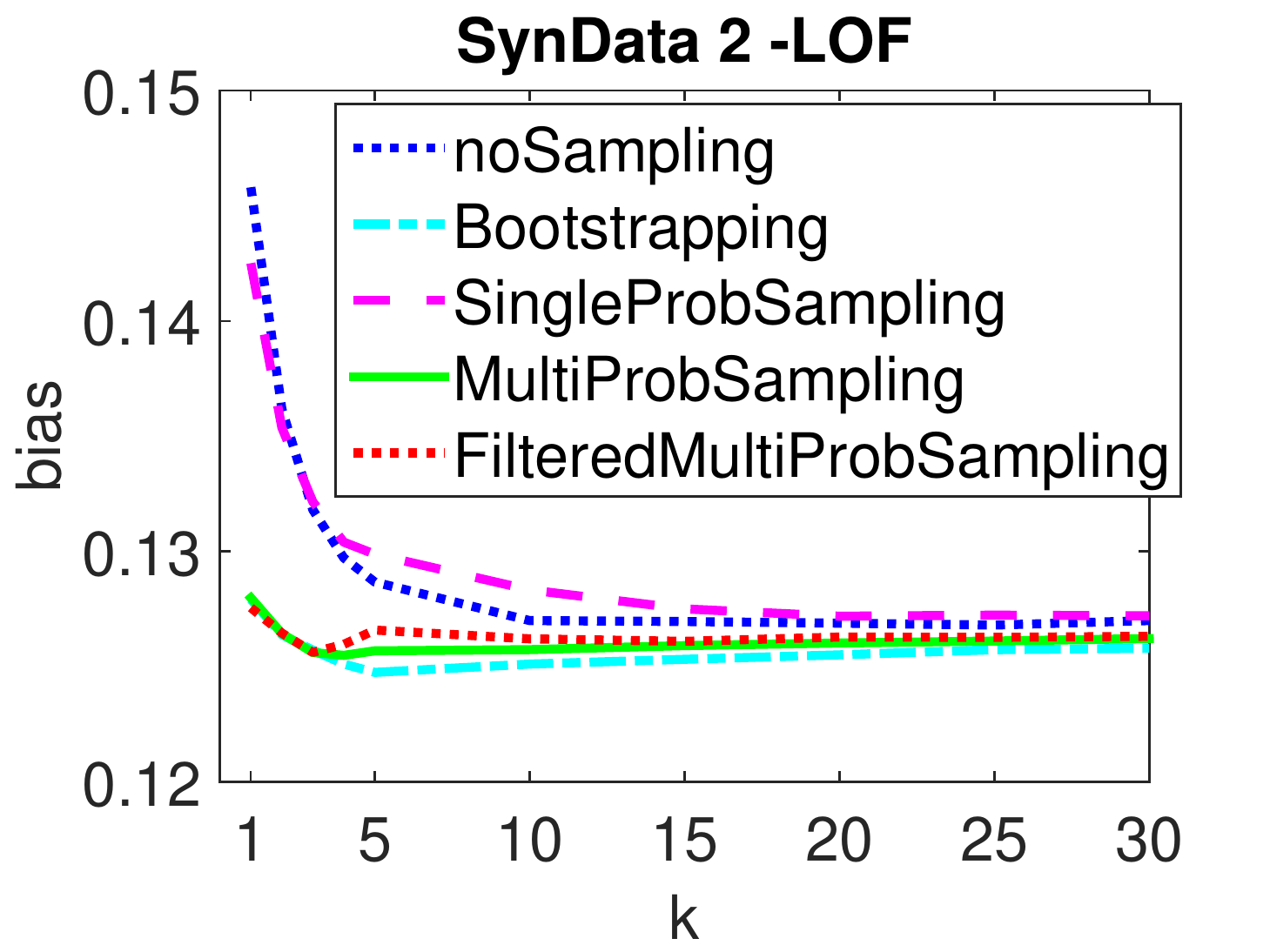}&
		\hspace{-0.2in}\includegraphics[width=0.51\linewidth,height=1.5in]{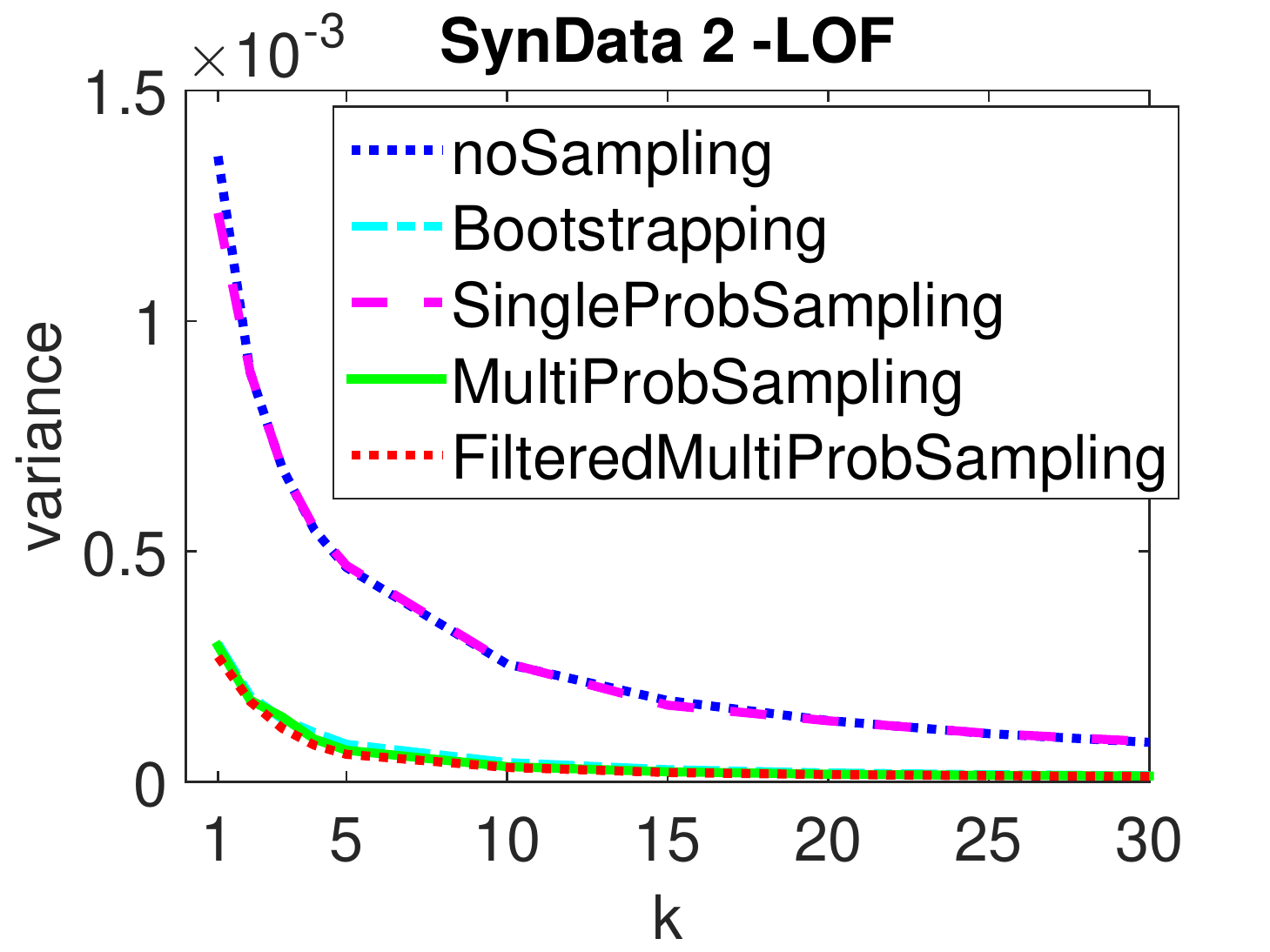}\\
		(c) & (d) \\
		\end{tabular}
	\vspace{-0.1in}
	\caption{bias (left) and variance (right)  vs. $k$ (avg'ed over 10 test datasets) on two synthetic datasets. Notice that our approach (red) w/ probability sampling after top outliers being filtered reduces both bias and variance.
		\label{fig:bv}}
	\vspace{-0.2in}
\end{figure}

According to~\cite{aggarwal2015theoretical}, ensembles with feature-bagged base detectors and with variable sampling tend to reduce variance. In this section, we provide quantitative results through experiments on synthetic datasets to show that filtering top $T$ outliers and probability sampling in our sequential ensemble reduce bias along with variance. To present the bias-variance reduction quantitatively, we design five procedures. 
For each synthetic dataset, we use a data generation model $\mathbb{M}$ to create $R$ training datasets $D_{i}, ~ i = 1 \ldots R$ of size $m = 210$ (200 inliers and 10 outliers) and 10 test datasets $D_{j}^{Test}, ~j = 1 \ldots 10$ of size $n=1000$ by randomly drawing points from $\mathbb{M}$. Bias and variance of different procedures for different values of $k$ (i.e., \# nearest neighbors) for a test data $D_{j}^{Test}$ are calculated w.r.t. the training data $D_{i}^{'}, ~i = 1 \ldots R$ sampled from $D_i$ as follows:
\begin{equation}
	bias = \sqrt{\frac{\sum_{x=1}^{n}{(f^{*}(x) - \overline{f}(x))^{2}}}{n}}
\end{equation}
\vspace{-0.05in}
\begin{equation}
	var = \frac{\sum_{x=1}^{n}{\sum_{i=1}^{R}{(f(x,D_{i}^{'},k) - \overline{f}(x))^{2}}}}{n \times R}
\end{equation}

\begin{figure*}[t!]
	\vspace{-0.1in}
	\centering
	\begin{tabular}{c}
		\hspace{-0.1in}\includegraphics[width=\linewidth,height=2.8in]{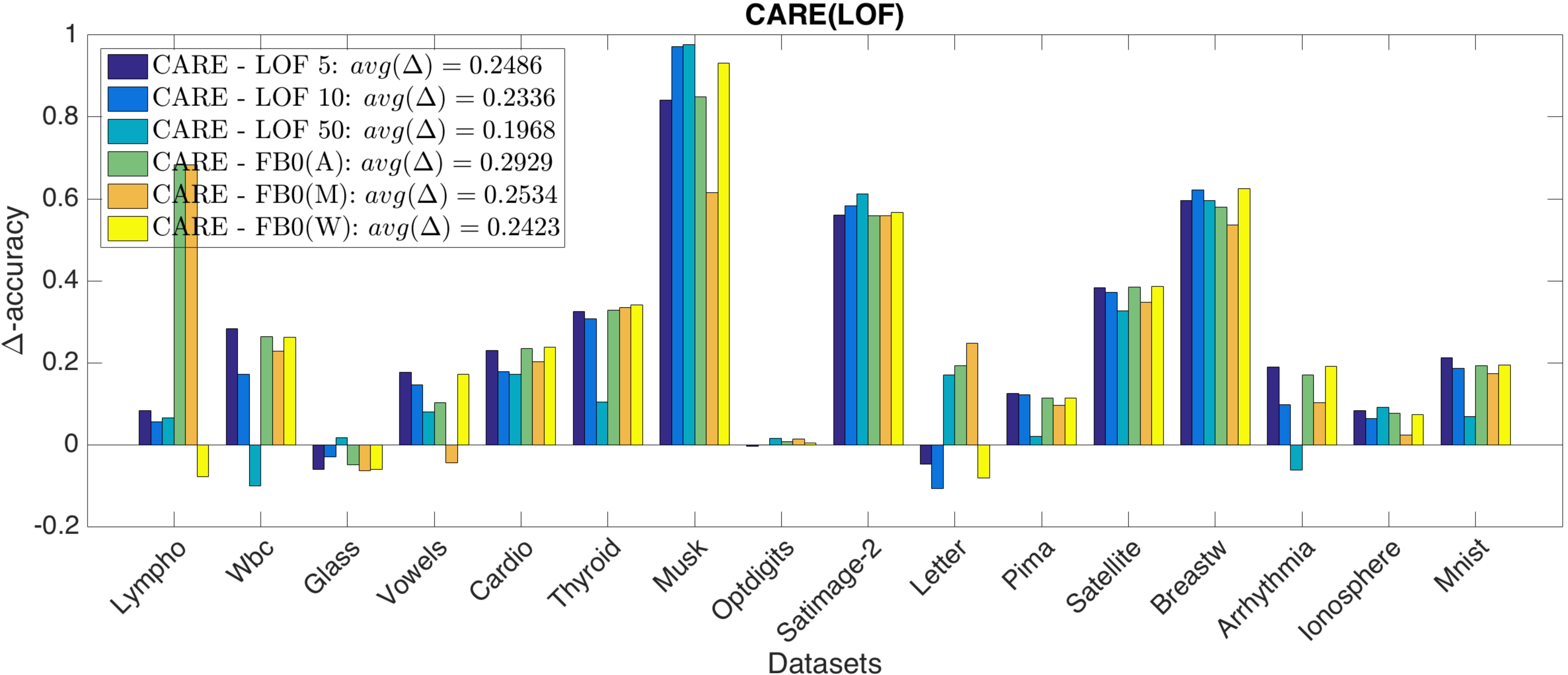}
		\end{tabular}
	\vspace{-0.1in}
	\caption{$\Delta$AP (Average Precision) from \scare(\slof)~to \slof~based baseline approaches on all the datasets. Notice that \scare~boosts detection performance significantly for  14/16 datasets  over most of the baseline approaches. $avg(\Delta)$ denotes average of $\Delta$AP values across datasets.
		\label{fig:delta1}}
	\vspace{-0.1in}
\end{figure*}

Here, $\overline{f}(x) = \frac{\sum_{i=1}^{R}f(x,D_{i}^{'},k)}{R}$, $f^{*}(x)$ is the actual label of data point $x \in D_{j}^{Test}$, and $f(x,D_{i}^{'},k)$ is the normalized outlierness score of $x$ w.r.t. sampled training set $D_{i}^{'}$ for $k$ nearest neighbors. For each procedure we design a different approach for sampling $D_{i}^{'}$. These five different procedures are briefly described as follows: \\
($i$) \textit{noSampling}: $D_{i}^{'} = D_{i}$, ($ii$) \textit{Bootstrapping}: sampling $m$ times (w/ replacement) from $D_{i}$ to get $D_{i}^{'}$, ($iii$) \textit{SingleProbSampling}: probability sampling on $f(D_i, D_i, k)$ for a single iteration to get $D_{i}^{'}$,  ($iv$) \textit{MultiProbSampling}:  probability sampling on $f(D_{i}, D_{i}^{'}, k)$ for multiple (i.e. 10) iterations  where $D_{i}^{'} = D_{i}$ initially, and ($v$) \textit{FilteredMultiProbSampling}: filtered (top $T$ outliers removed from $D_i$) probability sampling on $f(D_{i}, D_{i}^{'}, k)$ for multiple iterations (i.e. 10) where $D_{i}^{'} = D_{i}$ initially (our proposed approach).

In this section, we provide results on only two synthetic datasets (20 dimensional) for brevity, where the inliers are drawn from a mixture of Gaussian distributions and outliers are drawn from ($i$)  power law, and ($ii$) uniform distribution. Figure~\ref{fig:bv} shows bias (left) and variance (right) vs. $k$, where for the top two plots (i.e. (a), (b)) \avgknn~is used to calculate $f(x,D_{i}^{'},k)$, and for the bottom two plots (i.e. (c), (d)) \lof~is used. We can see from the figure that the curve for MultiProbSampling (green) is below the noSampling (blue) as well as the SingleProbSampling (magenta) curve, showing that probability sampling in multiple iterations helps to reduce both bias and variance. We also see that the FilteredMultiProbSampling (red) reduces bias further thanks to the filtering of top $T$ outliers. Moreover, removing top outliers appears to also reduce variance as the red curve is below all the others in both (b) and (d).

\vspace{-0.05in}

\section{Experiments}
\label{sec:eval}

\vspace{-0.1in}
\subsection{Datasets}
\vspace{-0.1in}
We evaluate \care~on 16 different real-world outlier detection datasets\footnote{\url{http://odds.cs.stonybrook.edu/}} mostly from the UCI ML repository~\cite{Lichman:2013}. Table~\ref{tab:data} provides the summary of the datasets used in this work. The first 9 datasets, Letter dataset, and the following 5 datasets are respectively obtained from \cite{aggarwal2015theoretical}, \cite{micenkova2014learning} and \cite{ting2009mass}. 

\begin{table}[h]
	\vspace{-0.05in}
	\caption{{Real-world datasets used for evaluation, where $d$ is data dimensionality, and \% indicates the \% of outliers}.\label{tab:data}}
	\vspace{-0.15in}
	\small{
		\begin{center}
			\begin{tabular}{|l|r|r|l|}
				\hline		
				\textbf{Dataset} & \textbf{\#Pts} $n$  & \textbf{Dim.} $d$ & $\%$ \textbf{Outlier Class}\\
				\hline \hline
				Lympho & 148 & 18 & classes 1,4 (4.1\%) \\
				\hline
				WBC & 278 & 30 & \pbox{7cm} {21 sampled malignant\\ class (5.6\%)} \\
				\hline
				Glass & 214 & 9 & class 6 (4.2\%) \\
				\hline
				Vowels & 1456 & 12 & \pbox{7cm} {50 sampled class 1 (3.4\%),\\ classes 6,7,8, inliers} \\
				\hline
				Cardio & 1831 & 21 & \pbox{7cm} {176 sampled pathologic\\ (9.6\%), normal inliers} \\
				\hline
				Thyroid & 3772 & 6 & from \cite{keller2012hics} (2.5\%)\\
				\hline
				Musk & 3062 & 166 & \pbox{7cm} {classes 213,211 (3.2\%) \\classes j146,j147,252\\ inliers} \\
				\hline
				Optdigits & 5216 & 64 & 150 sampled digit 0 (3\%)\\
				\hline
				Satimage-2 & 5803 & 36 & 71 sampled class 2 (1.2\%)\\
				\hline
				Letter & 1600 & 32 & from \cite{micenkova2014learning} (6.25\%)\\
				\hline
				Pima & 768 & 8 & pos class (35\%) \\
				\hline
				Satellite & 6435 & 36 & 3 smallest classes (32\%) \\
				\hline
				Breastw & 683 & 9 & malignant class (35\%)\\
				\hline
				Arrhythmia & 452 & 274 & \pbox{7cm} {classes 3,4,5,7,8,9,14,15 \\(15\%)}\\
				\hline
				Ionosphere & 351 & 33 & bad class (36\%)\\
				\hline
				Mnist & 7603 & 100 & \pbox{7cm} {700 sampled digit 6 \\(9.2\%), digit 0 inliers}\\
				\hline
			\end{tabular}
		\end{center}
	}
	\vspace{-0.2in}
\end{table}

\subsection{Results}
\label{ssec:results}
\subsubsection{CARE vs state-of-the-art baselines}
We first compare \care~with simple \lof~and \avgknn~based baseline approaches; using $k = \{5, 10, 50\}$, as well as non-sequential feature bagging (FB0) approaches with three types of aggregation; average (A), maximum (M), and  weighted (W). Figure~\ref{fig:delta1} shows the $\Delta$ Average Precision (AP: area under 
\begin{figure*}[th]
	\centering
	\begin{tabular}{c}
		\hspace{-0.1in}\includegraphics[width=\linewidth,height=2.6in]{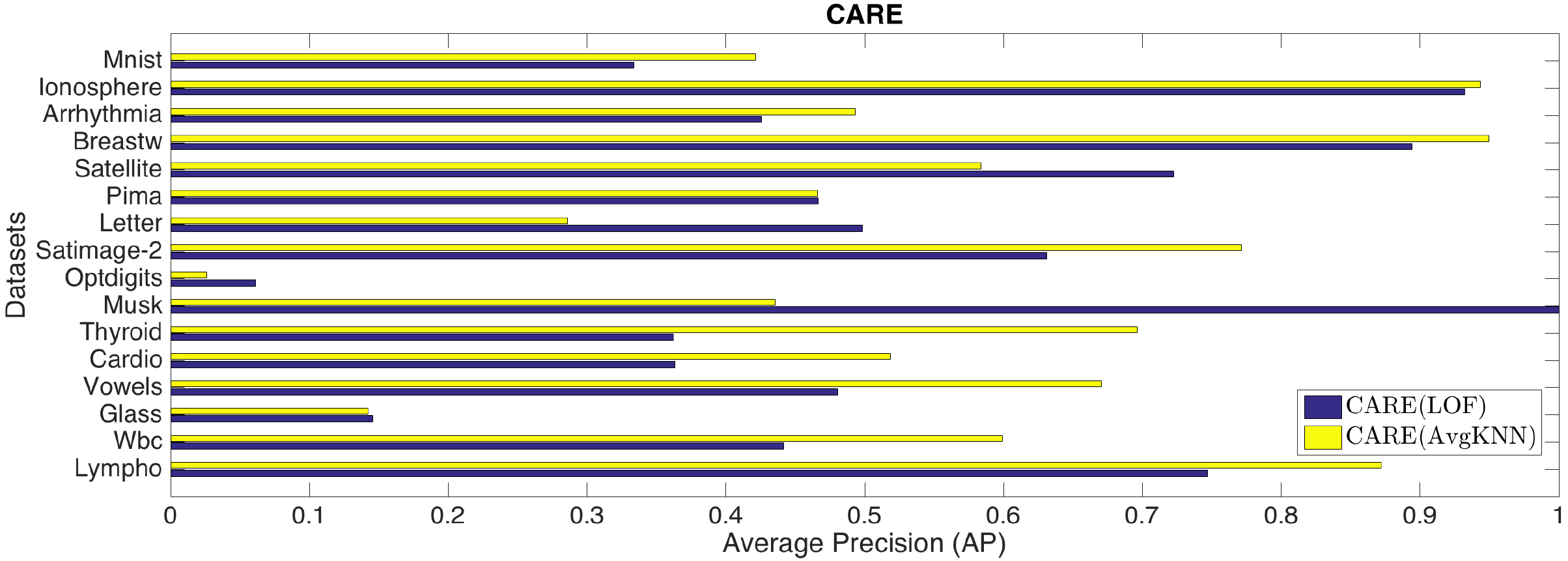}
		\end{tabular}
	\vspace{-0.1in}
	\caption{Average Precision (AP) of \scare~across datasets for both \slof~and \savgknn~based base detectors. \scare(\savgknn) performs better than \scare(\slof) on 10/16 datasets. 
		\label{fig:ap}}
\end{figure*}

\begin{figure*}[thb!p]
	\centering
	\begin{tabular}{c}
		\hspace{-0.1in}\includegraphics[width=\linewidth,height=2.8in]{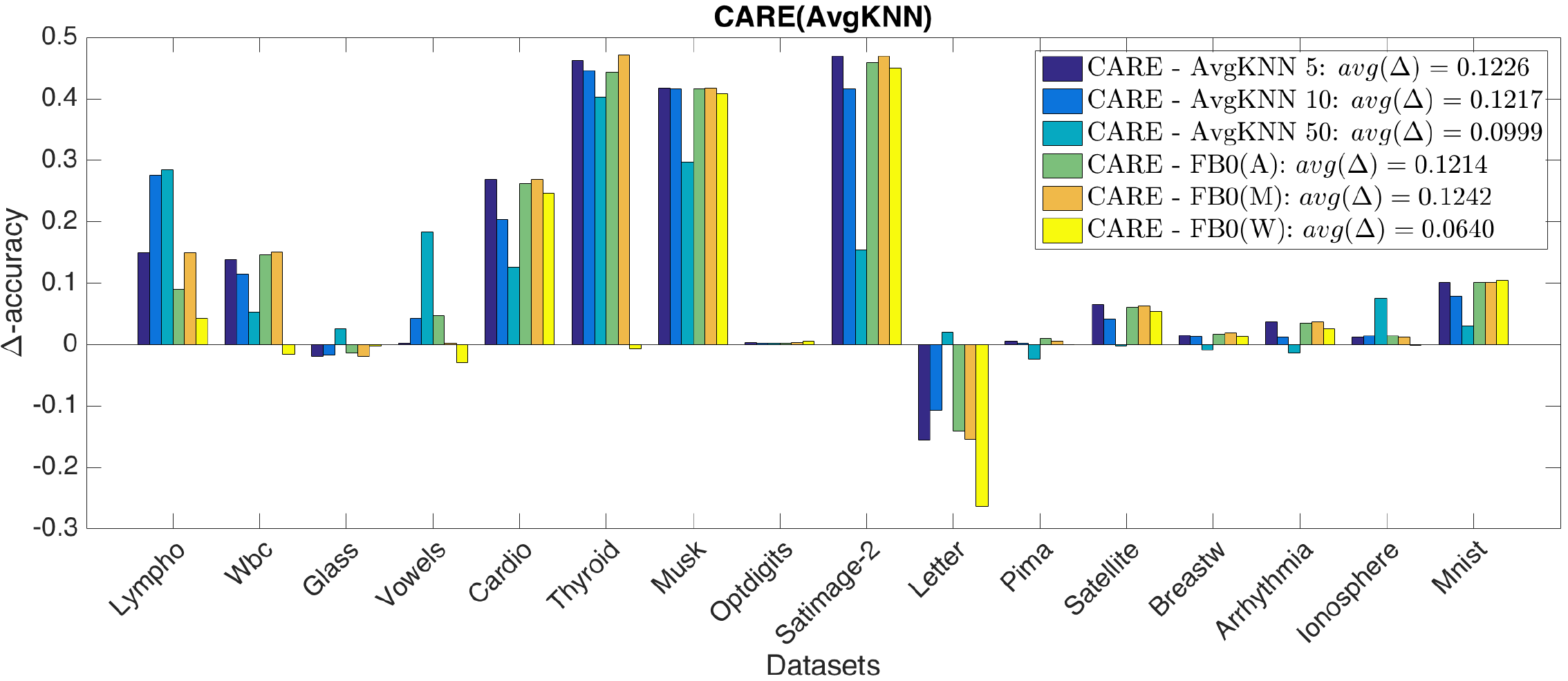}
		\end{tabular}
	\vspace{-0.1in}
	\caption{$\Delta$~AP values from \scare(\savgknn)~to \savgknn~based baselines. \scare~improves over more than half of the baselines on 14/16 datasets. 
		\label{fig:delta2}}
	\vspace{-0.1in}
\end{figure*}
\noindent
the precision-recall curve) values from \care(\lof) to these six state-of-the-art baselines all using the \lof~algorithm.
That is, the bars depict AP$^{\care}-$AP$^{\text{baseline}}$.
We refer to Figure \ref{fig:ap} for the original AP values that \care(\lof) and \care(\avgknn) achieve on the datasets.
Results show that  \care~outperforms all the base detectors on 9/16 datasets, and more than half of them on 14/16 datasets. Negative $\Delta$ values are much smaller as compared to positive ones, which indicates that in cases where \care~is not better than the baselines, it remains close. In the legend of the figure, we provide the overall $\Delta$AP values averaged across all the datasets and positive values indicate that \care~performs better than the individual baselines on average. Similarly, Figure~\ref{fig:delta2} contains the $\Delta$AP values from \care(\avgknn) to six baselines, which this time use \avgknn~based subroutines. Again, the average $\Delta$~values (in the legend) across different datasets indicate that \care~outperforms  the individual baselines on average. In cases where \care~falls shorter it often remains close to the baselines (note the relatively much smaller negative $\Delta$'s). From these two figures we also conclude that \care(\lof) provides greater improvement over the baselines compared to \care(\avgknn). 

\begin{figure*}[th!p]
	\vspace{-0.1in}
	\centering
	\begin{tabular}{c}
		\hspace{-0.1in}\includegraphics[width=\linewidth,height=2.8in]{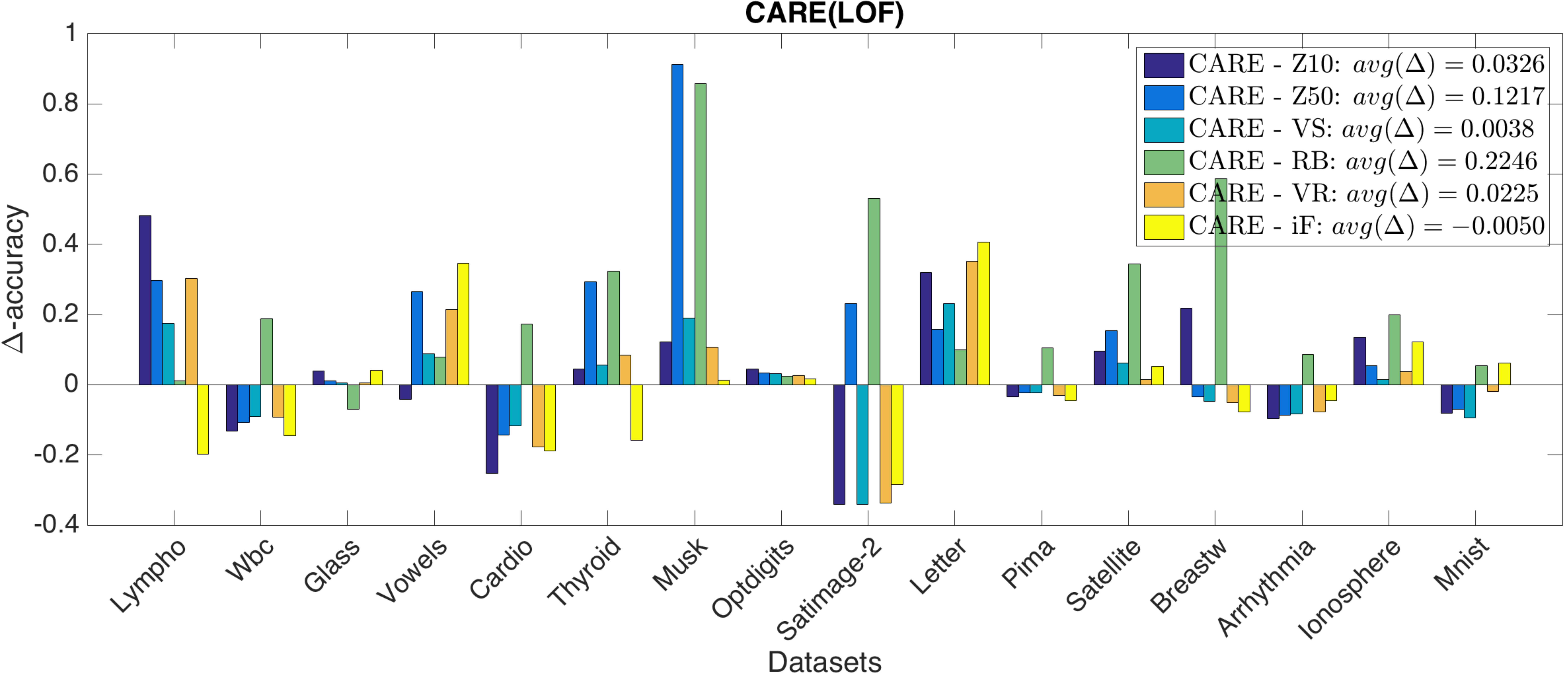}
		\end{tabular}
	\vspace{-0.1in}
	\caption{$\Delta$AP  from \scare(\slof)~to \slof~based state-of-the-art ensemble approaches on all the datasets. Notice that \scare~outperforms existing ensembles significantly on several datasets and achieves comparable performance otherwise. $avg(\Delta)$'s in the legend denote average of $\Delta$AP values across datasets.
		\label{fig:delta3}}
	\vspace{-0.1in}
\end{figure*}

\subsubsection{CARE vs state-of-the-art ensembles}
\begin{figure*}[th!p]
	\vspace{-0.05in}
	\centering
	\begin{tabular}{c}
		\hspace{-0.1in}\includegraphics[width=\linewidth,height=2.8in]{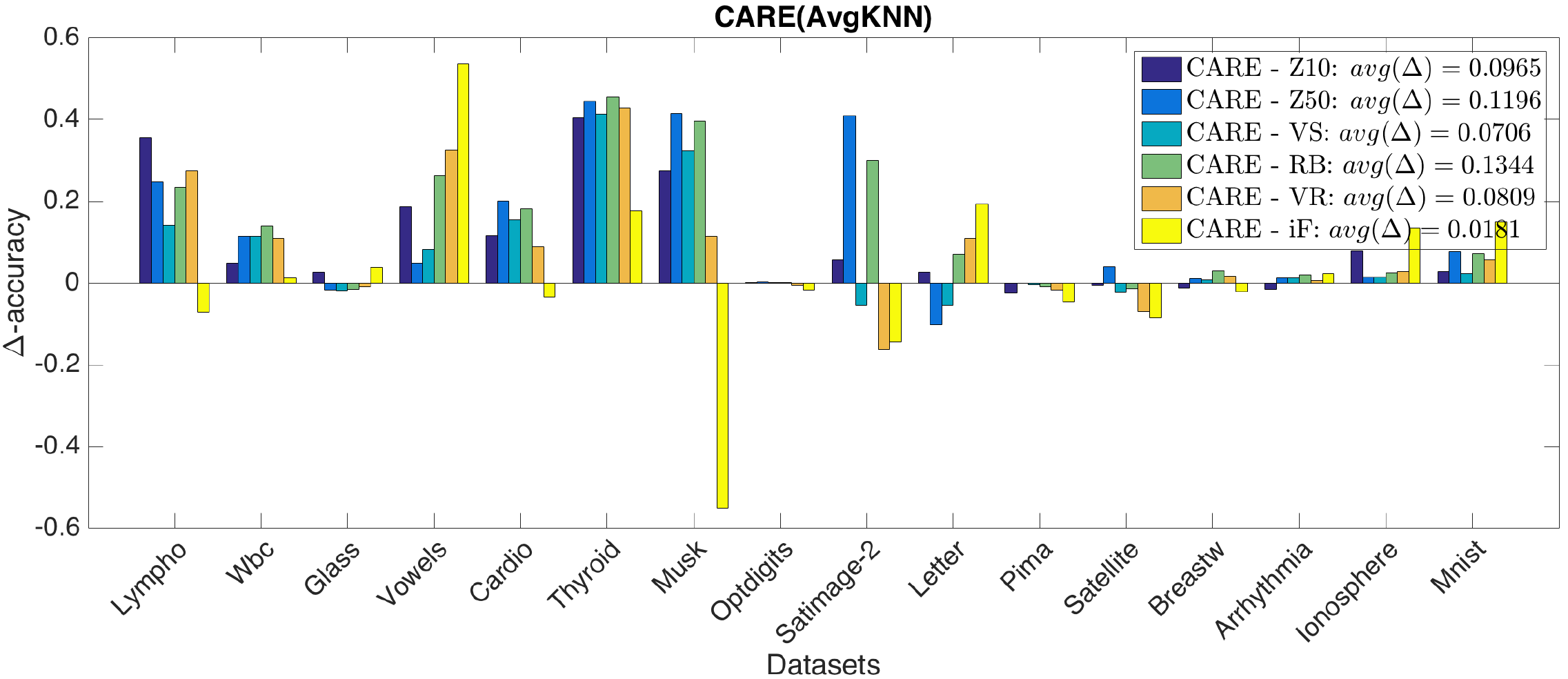}
		\end{tabular}
	\vspace{-0.1in}
	\caption{$\Delta$~AP values from \scare(\savgknn)~to \savgknn~based state-of-the-art ensemble approaches. Note the generally large positive and otherwise small negative values across datasets.
		\label{fig:delta4}}
	\vspace{-0.1in}
\end{figure*}

Next we compare \care~with the existing state-of-the-art outlier ensemble methods, including Aggarwal and Sathe's variable sampling (VS), rotated bagging (RB), and variable rotated bagging (VR) approaches~\cite{aggarwal2015theoretical}, Zimek \etal's subsampling approach~\cite{zimek2013subsampling}, as well as the Isolation Forest  (iF) ensemble of Liu \etal~\cite{liu2008isolation}. 
We employ $b = 100$ base detectors for each of these existing ensemble approaches such that they are comparable with \care. 
We present the $\Delta$AP from \care(\lof) to these six existing state-of-the-art outlier ensembles using the \lof~algorithm (except for iF) in Figure~\ref{fig:delta3}. For Zimek's subsampling approach we only present the results for sample sizes $10\%$ and $50\%$ (Z10, Z50). In Figure~\ref{fig:delta3}, we can see that the performance of \care(\lof) and VS are close with $avg(\Delta) = 0.0038$ across all the datasets. Notice that \care~mostly improves over Z50 and RB. Although iF is little better than \care(\lof) with $avg(\Delta) = -0.0050$, for some datasets e.g., Vowels and Letter where iF performs poorly with AP values 0.1341 and 0.0929 respectively, \care(\lof) provides $2.6\times$ improvement with AP value 0.4803 for Vowels, and $4.4\times$ improvement with AP value 0.4986 for Letter. Moreover, we note that  the magnitude of positive $\Delta$ values are larger than the negative ones on average.
 This indicates that \care(\lof) provides major improvement in cases when it is the winner and performs similarly to existing ensembles in other cases. Finally, Figure~\ref{fig:delta4} shows the corresponding results  for \care(\avgknn). Positive average $\Delta$AP values across all datasets show that \care~provides significant improvement when it outperforms an existing ensemble and falls short by a small margin in other cases.
 iF outperforms \care~significantly on Musk, which has a dense cluster of outliers that avoid detection by nearest neighbor based methods.
 We also find that none of the existing methods on Optdigits and Glass, where further investigation is needed to understand the type of outliers that they exhibit.

\section{Conclusion}
\label{sec:conclude}

\vspace{-0.05in}
In this paper, we proposed \care, a new sequential ensemble approach for outlier mining with a goal to achieve low detection error through reduced variance and bias.
Two main components of \care~are its parallel and sequential building blocks.
The former helps reduce variance by a weighted combination of multiple base detectors. Detector weights are derived from their error rates that are estimated through their relation to pairwise agreement rates.
On the other hand, the sequential component is designed to reduce bias. It utilizes results from previous iterations and a new sampling strategy \fvps~to weed out top outliers so as to construct a more robust data model based on which outlierness scores are computed.
We evaluate our method on 16 real-world datasets. Extensive experiments validate that \care~provides significant improvement over the baseline methods as well as the state-of-the-art outlier ensembles when it is the winner and performs close enough otherwise.  All source codes of our methods are shared openly at \url{http://shebuti.com/sequential-ensemble-learning-for-outlier-detection/}.

 \section*{Acknowledgment}
{This material is based upon work supported by the ARO Young Investigator Program under Contract No. W911NF-14-1-0029, NSF CAREER 1452425, IIS 1408287 and IIP1069147, DARPA Transparent Computing Program under Contract No. FA8650-15-C-7561, a Facebook Faculty Gift, an R\&D grant from Northrop Grumman Aerospace Systems, and Stony Brook University Office of Vice President for Research. Any conclusions expressed in this material are of the authors' and do not necessarily reflect the views, either expressed or implied, of the funding parties.}

\balance
\footnotesize{

}


\begin{thebibliography}{10}
\providecommand{\url}[1]{#1}
\csname url@samestyle\endcsname
\providecommand{\newblock}{\relax}
\providecommand{\bibinfo}[2]{#2}
\providecommand{\BIBentrySTDinterwordspacing}{\spaceskip=0pt\relax}
\providecommand{\BIBentryALTinterwordstretchfactor}{4}
\providecommand{\BIBentryALTinterwordspacing}{\spaceskip=\fontdimen2\font plus
\BIBentryALTinterwordstretchfactor\fontdimen3\font minus
  \fontdimen4\font\relax}
\providecommand{\BIBforeignlanguage}[2]{{%
\expandafter\ifx\csname l@#1\endcsname\relax
\typeout{** WARNING: IEEEtran.bst: No hyphenation pattern has been}%
\typeout{** loaded for the language `#1'. Using the pattern for}%
\typeout{** the default language instead.}%
\else
\language=\csname l@#1\endcsname
\fi
#2}}
\providecommand{\BIBdecl}{\relax}
\BIBdecl

\bibitem{aggarwal2015theoretical}
C.~C. Aggarwal and S.~Sathe, ``Theoretical foundations and algorithms for
  outlier ensembles.'' \emph{ACM SIGKDD Explorations Newsletter}, vol.~17,
  no.~1, pp. 24--47, 2015.

\bibitem{Lichman:2013}
\BIBentryALTinterwordspacing
M.~Lichman, ``{UCI} machine learning repository,'' 2013. [Online]. Available:
  \url{http://archive.ics.uci.edu/ml}
\BIBentrySTDinterwordspacing

\bibitem{breunig2000lof}
M.~M. Breunig, H.-P. Kriegel, R.~T. Ng, and J.~Sander, ``Lof: identifying
  density-based local outliers,'' in \emph{ACM sigmod record}, vol.~29,
  no.~2.\hskip 1em plus 0.5em minus 0.4em\relax ACM, 2000, pp. 93--104.

\bibitem{kriegel2009loop}
H.-P. Kriegel, P.~Kr{\"o}ger, E.~Schubert, and A.~Zimek, ``Loop: local outlier
  probabilities,'' in \emph{CIKM}.\hskip 1em plus 0.5em minus 0.4em\relax ACM,
  2009, pp. 1649--1652.

\bibitem{papadimitriou2003loci}
S.~Papadimitriou, H.~Kitagawa, P.~B. Gibbons, and C.~Faloutsos, ``Loci: Fast
  outlier detection using the local correlation integral,'' in
  \emph{ICDE}.\hskip 1em plus 0.5em minus 0.4em\relax IEEE, 2003, pp. 315--326.

\bibitem{zhang2009new}
K.~Zhang, M.~Hutter, and H.~Jin, ``A new local distance-based outlier detection
  approach for scattered real-world data,'' in \emph{Advances in Knowledge
  Discovery and Data Mining}, 2009, pp. 813--822.

\bibitem{otey2006fast}
M.~E. Otey, A.~Ghoting, and S.~Parthasarathy, ``Fast distributed outlier
  detection in mixed-attribute data sets,'' \emph{Data Mining and Knowledge
  Discovery}, vol.~12, no. 2-3, pp. 203--228, 2006.

\bibitem{zimek2013subsampling}
A.~Zimek, M.~Gaudet, R.~J. Campello, and J.~Sander, ``Subsampling for efficient
  and effective unsupervised outlier detection ensembles,'' in \emph{ACM
  SIGKDD}, 2013, pp. 428--436.

\bibitem{rayana2015less}
S.~Rayana and L.~Akoglu, ``Less is more: Building selective anomaly ensembles
  with application to event detection in temporal graphs.'' \emph{SDM},
  vol.~17, 2015.

\bibitem{brei96}
L.~Breiman, ``Bagging predictors,'' \emph{Machine Learning}, vol.~24, no.~2,
  pp. 123--140, 1996.

\bibitem{freund1997decision}
Y.~Freund and R.~E. Schapire, ``A decision-theoretic generalization of on-line
  learning and an application to boosting,'' \emph{Journal of computer and
  system sciences}, vol.~55, no.~1, pp. 119--139, 1997.

\bibitem{lazarevic2005feature}
A.~Lazarevic and V.~Kumar, ``Feature bagging for outlier detection,'' in
  \emph{ACM SIGKDD}, 2005, pp. 157--166.

\bibitem{gao2006converting}
J.~Gao and P.-N. Tan, ``Converting output scores from outlier detection
  algorithms into probability estimates,'' in \emph{ICDM}, 2006, pp. 212--221.

\bibitem{journals/sigkdd/Aggarwal12}
C.~C. Aggarwal, ``Outlier ensembles: position paper.'' \emph{SIGKDD Explor.
  Newsl.}, vol.~14, no.~2, pp. 49--58, 2012.

\bibitem{platanios2014estimating}
A.~Platanios, A.~Blum, and T.~M. Mitchell, ``Estimating accuracy from unlabeled
  data,'' in \emph{In Proceedings of UAI}, 2014.

\bibitem{liu2008isolation}
F.~T. Liu, K.~M. Ting, and Z.-H. Zhou, ``Isolation forest,'' in
  \emph{ICDM}.\hskip 1em plus 0.5em minus 0.4em\relax IEEE, 2008, pp. 413--422.

\bibitem{breiman1999using}
L.~Breiman, ``Using adaptive bagging to debias regressions,'' Statistics Dept.
  UCB, Tech. Rep., 1999.

\bibitem{sun2007cost}
Y.~Sun, M.~S. Kamel, A.~K. Wong, and Y.~Wang, ``Cost-sensitive boosting for
  classification of imbalanced data,'' \emph{Pattern Recognition}, vol.~40,
  no.~12, pp. 3358--3378, 2007.

\bibitem{breiman2001random}
L.~Breiman, ``Random forests,'' \emph{Machine Learning}, vol.~45, pp. 5--32,
  2001.

\bibitem{aggarwal2013outlier}
C.~C. Aggarwal, ``Outlier ensembles: position paper,'' \emph{ACM SIGKDD
  Explorations Newsletter}, vol.~14, no.~2, pp. 49--58, 2013.

\bibitem{Zimek13Ensemble}
A.~Zimek, R.~J. Campello, and J.~Sander, ``Ensembles for unsupervised outlier
  detection: Challenges and research questions,'' \emph{SIGKDD Explor. Newsl.},
  vol.~15, no.~1, pp. 11--22, 2013.

\bibitem{hawkins1980identification}
D.~M. Hawkins, \emph{Identification of outliers}.\hskip 1em plus 0.5em minus
  0.4em\relax Springer, 1980, vol.~11.

\bibitem{knorr1997unified}
E.~M. Knorr and R.~T. Ng, ``A unified notion of outliers: Properties and
  computation.'' in \emph{KDD}, 1997, pp. 219--222.

\bibitem{RayanaAkoglu14}
S.~Rayana and L.~Akoglu, ``An ensemble approach for event detection in dynamic
  graphs.'' in \emph{ACM SIGKDD ODD$^2$ Workshop}, 2014.

\bibitem{grimmet2001}
G.~Grimmett and D.~Stirzaker, \emph{Probability and Random Processes.}, 2001.

\bibitem{Zimek11Uni}
H.-P. Kriegel, P.~Kr\"{o}ger, E.~Schubert, and A.~Zimek, ``Interpreting and
  unifying outlier scores.'' in \emph{SDM}, 2011.

\bibitem{micenkova2014learning}
B.~Micenkov{\'a}, B.~McWilliams, and I.~Assent, ``Learning outlier ensembles:
  The best of both worlds--supervised and unsupervised,'' in \emph{ACM SIGKDD
  ODD$^2$ Workshop}, 2014.

\bibitem{ting2009mass}
K.~Ting, S.~Tan, and F.~Liu, ``Mass: A new ranking measure for anomaly
  detection,'' \emph{Gippsland School of Information Technology, Monash
  University}, 2009.

\bibitem{keller2012hics}
F.~Keller, E.~M{\"u}ller, and K.~B{\"o}hm, ``Hics: high contrast subspaces for
  density-based outlier ranking,'' in \emph{Data Engineering (ICDE), 2012 IEEE
  28th International Conference on}.\hskip 1em plus 0.5em minus 0.4em\relax
  IEEE, 2012, pp. 1037--1048.

\end{thebibliography}
\end{document}